\documentclass{article}

\usepackage{microtype}
\usepackage{graphicx, subfig}
\usepackage{booktabs} 
\usepackage{enumitem}
\usepackage{hyperref}

\usepackage[accepted]{icml2019}

\usepackage{color,xcolor}
\usepackage{epsfig}
\usepackage{graphicx}

\usepackage{adjustbox}
\usepackage{array}
\usepackage{booktabs}
\usepackage{colortbl}
\usepackage{float,wrapfig}
\usepackage{hhline}
\usepackage{multirow}

\usepackage{amsmath,amsfonts,amsthm,amssymb}
\usepackage{bm}
\usepackage{nicefrac}
\usepackage{microtype}

\usepackage{changepage}
\usepackage{extramarks}
\usepackage{fancyhdr}
\usepackage{lastpage}
\usepackage{setspace}
\usepackage{soul}
\usepackage{xspace}

\usepackage{url}

\usepackage{algorithm, algorithmic}
\usepackage{todonotes} 

\usepackage{titlesec}

\newcolumntype{L}[1]{>{\raggedright\let\newline\\\arraybackslash\hspace{0pt}}m{#1}}
\newcolumntype{C}[1]{>{\centering\let\newline\\\arraybackslash\hspace{0pt}}m{#1}}
\newcolumntype{R}[1]{>{\raggedleft\let\newline\\\arraybackslash\hspace{0pt}}m{#1}}

\newcommand{\fig}[1]{Figure~\ref{#1}}

\newcommand{\ignore}[1]{}

\makeatletter
\DeclareRobustCommand\onedot{\futurelet\@let@token\@onedot}
\def\@onedot{\ifx\@let@token.\else.\null\fi\xspace}

\makeatother

\definecolor{MyDarkBlue}{rgb}{0,0.08,1}
\definecolor{MyDarkGreen}{rgb}{0.02,0.6,0.02}
\definecolor{MyDarkRed}{rgb}{0.8,0.02,0.02}
\definecolor{MyDarkOrange}{rgb}{0.40,0.2,0.02}
\definecolor{MyPurple}{RGB}{111,0,255}
\definecolor{MyRed}{rgb}{1.0,0.0,0.0}
\definecolor{MyGold}{rgb}{0.75,0.6,0.12}
\definecolor{MyDarkgray}{rgb}{0.66, 0.66, 0.66}

\newcommand{\myparagraph}[1]{\vspace{-5pt}\paragraph{#1}}

\titlespacing*{\section}{0pt}{0pt plus 1pt minus 0pt}{0pt plus 1pt minus 0pt}
\titlespacing\subsection{0pt}{0pt plus 0pt minus 0pt}{0pt plus 0pt minus 0pt}
\newcommand{\spatial}{SpatialNet\xspace}
\newcommand{\ohnet}{RCNet\xspace}
\newcommand{\clstm}{ConvLSTM\xspace}
\newcommand{\agent}{IPA\xspace}

\icmltitlerunning{Task-Agnostic Dynamics Priors for Deep Reinforcement Learning}

\begin{document}

\twocolumn[
\icmltitle{Task-Agnostic Dynamics Priors for Deep Reinforcement Learning}



\icmlsetsymbol{equal}{*}

\begin{icmlauthorlist}
\icmlauthor{Yilun Du}{mit}
\icmlauthor{Karthik Narasimhan}{pri}
\end{icmlauthorlist}

\icmlaffiliation{mit}{Massachusetts Institute of Technology (Work partially done at OpenAI)}
\icmlaffiliation{pri}{Princeton University}
\icmlcorrespondingauthor{Yilun Du}{yilundu@mit.com}
\icmlcorrespondingauthor{Karthik Narasimhan}{karthikn@cs.princeton.edu}

\icmlkeywords{Reinforcement Learning}

\vskip 0.3in
]



\printAffiliationsAndNotice{}  

\begin{abstract}
While model-based deep reinforcement learning (RL) holds great promise for sample efficiency and generalization, learning an accurate dynamics model is often challenging and requires substantial interaction with the environment. A wide variety of domains have dynamics that share common foundations like the laws of classical mechanics, which are rarely exploited by existing algorithms. In fact, humans continuously acquire and use such \emph{dynamics priors} to easily adapt to operating in new environments. In this work, we propose an approach to learn task-agnostic dynamics priors from videos and incorporate them into an RL agent. Our method involves pre-training a frame predictor on task-agnostic physics videos to initialize dynamics models (and fine-tune them) for unseen target environments. Our frame prediction architecture, \spatial, is designed specifically to capture localized physical phenomena and interactions. Our approach allows for both faster policy learning and convergence to better policies, outperforming competitive approaches on several different environments. We also demonstrate that incorporating this prior allows for more effective transfer between environments.\footnote{Code is available at \url{https://github.com/yilundu/task_agnostic_dynamics_prior}}


\end{abstract}

\section{Introduction}

Recent advances in deep reinforcement learning (RL) have largely relied  on model-free approaches, demonstrating strong performance on a variety of domains~\citep{Silver2016Mastering, Mnih2013Playing, kempka2016vizdoom,dota}. However, model-free techniques do not have good sample efficiency~\citep{Sutton1990Integrated} and are difficult to adapt to new tasks or domains~\citep{nichol2018gotta}. A key reason for this is  a single value function is used to represent both the agent's policy and its knowledge of environment dynamics, which can result in heavy overfitting to a particular task~\cite{zhang2018study}. On the other hand, model-based RL allows for decoupling the dynamics model from the policy, enabling better generalization and transfer across tasks~\citep{zhang2018decoupling}. The challenge with model-based RL, however, lies in estimating an accurate dynamics model of the environment while simultaneously using it to learn a policy, often leading to sub-optimal policies and slower learning. One way to alleviate this problem is to initialize dynamics models with universal \emph{task-agnostic} priors that allow for more efficient and stable model-based learning. 

\begin{figure}[t]
\centering
\includegraphics[width=\linewidth]{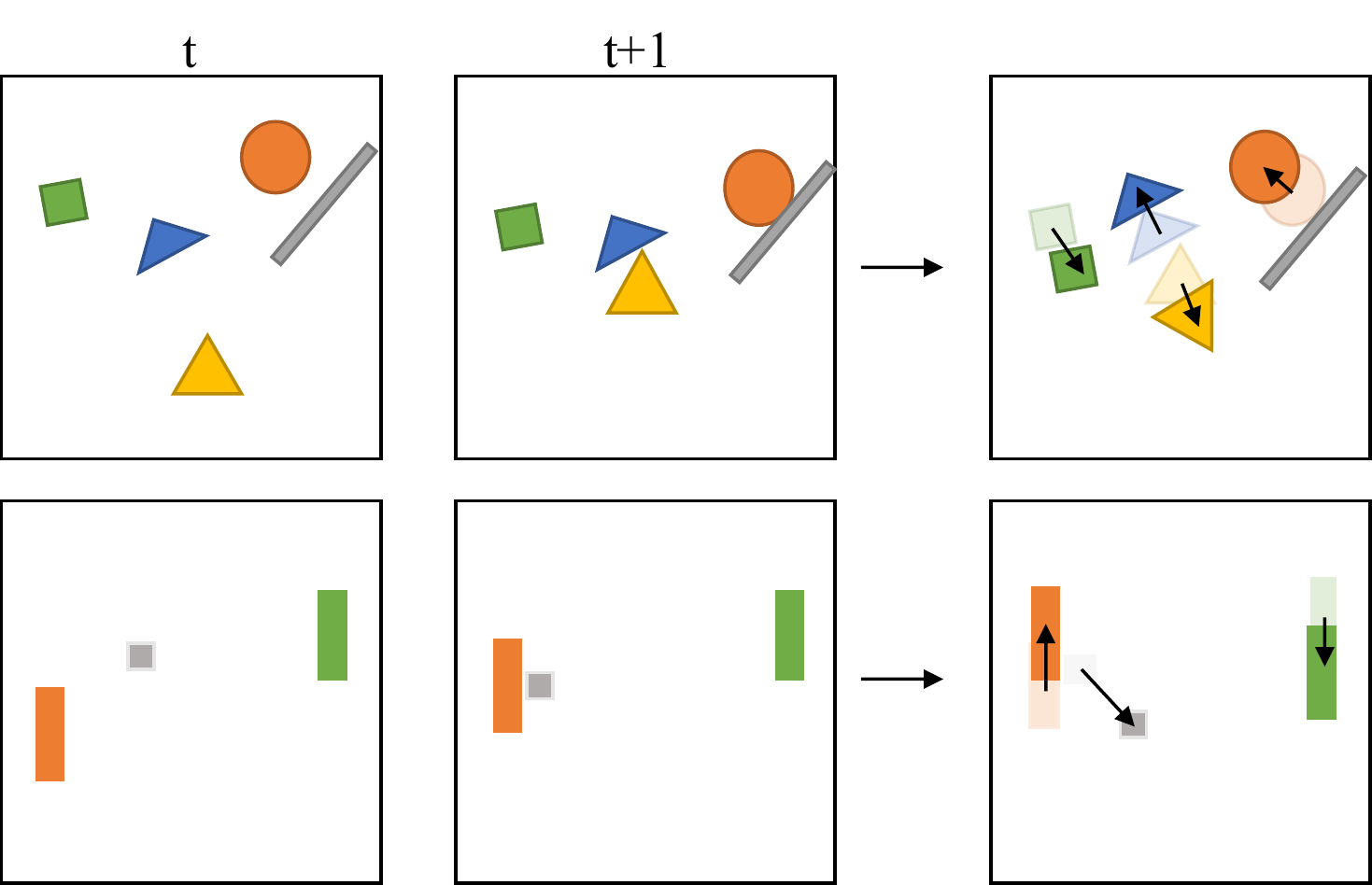}
\caption{\small Two different environments with object dynamics that obey the common laws of physics (\textit{top:} PhysWorld, \textit{bottom:} Atari Pong). Agents that have a knowledge of general physics will be able to adapt quickly to either environment.}
\label{fig:teaser}
\vspace{-20pt}
\end{figure}

For example, consider learning dynamics models for the two different scenarios shown in Figure~\ref{fig:teaser} (top and bottom). Both environments contain a variety of objects moving with different velocities and rotations. Current approaches require a large number of samples to learn a robust transition model of either world. For instance, in the first environment, inferring that the \emph{orange circle} is a freely moving object will require observing the circle moving in a variety of different directions. Or to understand the laws governing elastic collisions between two bodies (e.g. the circle and the grey rectangle) requires observing several instances of collisions at various angles and velocities. On the other hand, humans have reliable priors that allow for understanding dynamics of new environments  quickly~\citep{dubey2018investigating} -- 
one such prior is an understanding of physical laws of motion. In this work, we demonstrate that learning a task-agnostic dynamics prior (e.g. concepts like velocity, acceleration or elasticity) allows for accurate and more efficient estimation of the dynamics of new environments, resulting in better control policies. 


In order to obtain a prior for physical dynamics, we perform unsupervised learning over raw videos containing moving objects. Specifically, we train a dynamics model to predict the next frame given the previous k frames, over a wide variety of scenarios with moving objects. The parameters of the model implicitly capture general laws of physics, which are useful in predicting entity movements. We initialize the dynamics model of the environment with these pre-trained parameters and fine-tune them using transitions from the specific task, while simultaneously learning a policy for the task. The dynamics model is used to predict future frames up to a finite horizon, which are then used as additional input into a policy network, similar to the approach of \cite{weber2017imagination}. Importantly, our frame prediction model is not action-conditional like most prior work that employs such models in reinforcement learning~\cite{Oh2015Action,weber2017imagination}.

Learning a good future frame model is challenging mainly for two reasons: a) the large dimensionality of the output space with arbitrary moving objects and interactions, and b) the partial observability in environments~\citep{mathieu2015deep}. Prior  approaches~\citep{Oh2015Action} suffered from error compounding since they encode the entire image into a single vector before decoding the output, thereby missing out fine-grained spatial information. Others like the ConvLSTM~\citep{xingjian2015convolutional} are better at capturing spatio-temporal interactions but suffer from poor generalization due the use of additive update equations. To overcome these issues, we propose a new architecture (\spatial) that consists of a convolutional encoder, a spatial memory block, and a convolutional decoder that better captures localized dynamics. The spatial memory module  operates by performing convolution operations over a temporal 3-dimensional state representation that keeps spatial information intact.  This allows the network, which includes residual connections, to capture localized physics of objects such as directional movements and collisions in a fine-grained manner as well as efficiently keep track of static background information. This results in lower prediction error, better generalization and invariance to the size of inputs. 

We evaluate our approach on three different RL scenarios. First, we consider PhysWorld, a suite of randomized 2D physics-focused games, where learning object movement is crucial to a successful policy. Next we consider PhysShooter3D, a 3D environment with rigid body dynamics and partial observations. Finally, we also evaluate on a stochastic variant of the popular ALE framework consisting of Atari games~\cite{machado2017ale}. In all scenarios, we first demonstrate the value of learning a task-agnostic prior for model dynamics -— for instance, our agent achieves up to 130\% higher performance on a shooting game, PhysShooter and 56.5\% higher on the Atari game of Asteroids, compared to the most competitive baseline. Further, we also show that the dynamics model fine-tuned on these tasks transfers better to new tasks. For instance, our model achieves a relative score improvement of 26.9\% on transfer from PhysForage to PhysShooter (both games from PhysWorld), significantly higher than a score improvement of 5.4\% using a \emph{policy-transfer} baseline.



\section{Related Work}
There are two main lines of work that are closely related to this paper. The first is that of learning and using generic video prediction models for reinforcement learning~\cite{Oh2015Action,Finn2016Unsupervised,weber2017imagination}. The key idea is to train a model to predict future frames on the target task and hallucinate additional trajectories that can help an agent learn faster. The second direction is to incorporate physics priors into parameterized dynamics models for future state prediction~\cite{nguyen2010using,Kansky2017Schema}. The former path requires only pixel inputs but does not generalize well across tasks. The latter has the potential to generalize but requires manual specification of priors. Our work aims to combine the best of both worlds -- learn a frame prediction model that is task-agnostic and captures an effective notion of physics to serve as a useful prior.

\textbf{Video prediction models. }
Our frame prediction model is closest in spirit to the ConvolutionalLSTM model which has been applied to several domains~\citep{xingjian2015convolutional,zhu2017multimodal,Ke2017ShortTermFO}. Similar architectures that incorporate differentiable memory modules~\citep{Patraucean2015SpatiotemporalVA} or relational intermediates~\citep{Watters2017Visual} have been proposed, with applications to deep RL~\cite{Parisotto2017NeuralMS}. While the \clstm model is reasonably effective at predicting future frames, the additive LSTM update equations are not well suited to capture localized physical interactions.\footnote{While the model theoretically can learn to ignore unnecessary operations, optimizing the parameters effectively is difficult because of a lack of proper inductive bias in the architecture.} Our architecture is simpler and more natural at capturing physical dynamics and entity movements -- this allows for better generalization as we demonstrate in our experiments. 


Several recent methods have also combined policy learning with future frame prediction in different ways. Action-conditioned frame prediction has been used to simulate trajectories for policy learning~\citep{Oh2015Action,Finn2016Unsupervised,weber2017imagination}. Predicted frames have also been used to incentivize exploration in agents, via hashing~\citep{yin2017hashing} or using the prediction error to provide intrinsic rewards~\citep{Pathak2017Curiosity}.
The main departure of our work from these papers is that we learn a frame prediction model that is not conditioned on actions, and from videos not related to a task, which allows us to employ the model on a variety of tasks.

\textbf{Parameterized physics models. }
Several recent papers have explored the idea of incorporating physics priors into learning dynamics models of environments~\cite{nguyen2010using,cutler2014reinforcement,cutler2015efficient,scholz2014physics,Kansky2017Schema,Battaglia2016Interaction, mrowca2018flexible,xie2016model}.
More recent work trained an object-oriented dynamics predictor by segmenting input frames into sets of objects~\cite{NIPS2018_8187}. While all these approaches demonstrate the importance of having relevant priors to sample efficient model learning, they all require some form of manual parameterization. In contrast, we learn physics priors in the form of the parameters of a predictive neural network, only using raw videos.


\textbf{Decoupling dynamics from policy. }
Our work also relates to previous approaches on decoupling the agent's knowledge of the environment dynamics from its task-oriented policy. Successor representations~\cite{Dayan1993Improving} decompose the agent's value function into a feature-based state representation and a reward projection operator, resulting in better exploration of the state space~\cite{Kulkarni2016Deep,barreto2017successor,machado2017eigenoption}. While these state abstractions help with exploration, such representations do not explicitly capture dynamics models of the environment.  More recent work has proposed approaches to learn separate models for dynamics and rewards and use it to perform online planning~\cite{zhang2018decoupling} or learn independently controllable factors in the environment~\cite{thomas2017independently}. However, these assume access to task-specific transitions, while we learn a prior from task-independent videos and demonstrate its usefulness in learning different environment dynamics.

\section{Framework}
\label{sec:spatial}
Our goal is to demonstrate that acquiring task-agnostic dynamics priors from raw videos helps agents learn faster in new environments. To this end, our approach consists of two phases:
\begin{enumerate}
    \item \textbf{Pre-training a dynamics predictor}: We first train a suitable neural network architecture to predict pixels in the next frame given the previous $k$ frames of a video. In this work, we use videos of objects moving according to classical mechanics, without any extra annotations.
    \item \textbf{Reinforcement learning}: We use the pre-trained frame predictor from the previous phase to initialize the dynamics model for an RL agent. This dynamics model is used to predict a few frames into the future, which is used as additional context for the control policy. The dynamics model is also simultaneously fine-tuned using trajectories from the environment. 
\end{enumerate} 
We first describe how we use the frame prediction model for reinforcement learning, and then discuss different options for a frame predictor, including our new architecture, \spatial.

\subsection{Reinforcement Learning with Dynamics Predictors} \label{sec:p2a}
There are several ways one can incorporate a dynamics model into a reinforcement learning setup. One approach is to use the model to generate synthetic trajectories and use them in addition to observed transitions while training a policy~\citep{Oh2015Action, feinberg2018model}. Another option is to perform rollouts from the current step using the model and then use the predicted states as additional context input to the policy~\citep{weber2017imagination}. Our method is similar to the latter -- we use our learned dynamics model to predict $k$ future frames and concatenate these frames along with the current frame to form the input to our policy network. There are two differences however -- (1) we predict future state observations \textit{without conditioning on the actions of the agent} and without rewards since our dynamics model is task agnostic, and (2) we do not use a global encoding for future frames, but instead stack the frames and use convolution operations to extract local dynamic information.

Formally, consider a standard Markov Decision Process (MDP) setup represented by the tuple $\langle S, A, T, R\rangle$, where $S$ is the set of all possible state configurations, $A$ is the set of actions available to the agent, $T$ is the transition distribution, and $R$ is the reward function. Assuming our dynamics model to be $\Omega$, and given the current state $s_t$, we first apply our prediction model iteratively to obtain future state predictions:
\begin{equation*}
    \hat{s}_{t+1} = \Omega(s_t), \hat{s}_{t+2} = \Omega(\hat{s}_{t+1}), ~ ... ~  \hat{s}_{t+k} = \Omega(\hat{s}_{t+k-1})
\end{equation*}

We then train a policy network to output actions using all these predicted states as input in addition to the current state:
\begin{equation}
    a_t = \pi(s_t, \hat{s}_{t+1}, ~ ... ~ \hat{s}_{t+k})
\end{equation}
For the policy network, we follow the architecture described in \cite{Mnih2015Human} and use the Proximal Policy Optimization (PPO)~\citep{Schulman2017Proximal} algorithm for learning from rewards obtained in the task. 
We call this agent an Intuitive Physics Agent (\agent) since it first learns an intuitive prior of physical interactions.

We update policy parameters by using the PPO loss:
\begin{equation*}
    L(\theta) = \mathbb{E} [\min (r_t(\theta) A_t, \text{clip} (r_t(\theta), 1-\epsilon, 1+\epsilon)A_t] 
\end{equation*}
where $r_t = \tfrac{\pi_{\theta}(a_t|s_t, \hat{s}_{t+1}, ~ ... ~ \hat{s}_{t+k})}{\pi_{\theta_{\text{old}}}(a_t|s_t, \hat{s}_{t+1}, ~ ... ~ \hat{s}_{t+k})}$ and  the advantage, $A_t$, is computed using the value function $V(s_t, \hat{s}_{t+1}, ~ ... ~ \hat{s}_{t+k})$.
Simultaneously, we also update the parameters of the dynamics model using the  transitions from the environment with a pixel prediction loss (described in Section~\ref{sec:mseloss}.) 
However, policy gradients are not back-propagated to the dynamics predictor.



\subsection{Dynamics Prediction}
Prior work has investigated a variety of frame prediction models. LSTM-based recurrent networks~\citep{Oh2015Action} are not ideal for this task since they encode the entire scene into a single latent vector, thereby losing the localized spatio-temporal correlations that are important for making accurate physical predictions.
On the other hand, the \clstm~\citep{xingjian2015convolutional} architecture has localized spatio-temporal correlations, but is not able to accurately maintain global dynamics of entities due to LSTM state updates and limited separation of stationary and non-stationary objects. (as also seen in our experiments in Section~\ref{sec:physexp}). 

Predicting the physical behavior of an entity requires a model that can perform two crucial operations -- 1) isolation of the dynamics of each entity, and 2) accurate modeling of localized spaces and interactions around the entity. In order to satisfy both desiderata, we propose a new architecture, \spatial, which uses a spatial memory that explicitly encodes dynamics that are updated with object movement through convolutions. This allows us to implicitly capture and maintain localized physics, such as entity velocities and collisions between entities, in our frame prediction model and results in significantly lower long term prediction error. 

\begin{figure}[t]
\centering
\includegraphics[width=\linewidth]{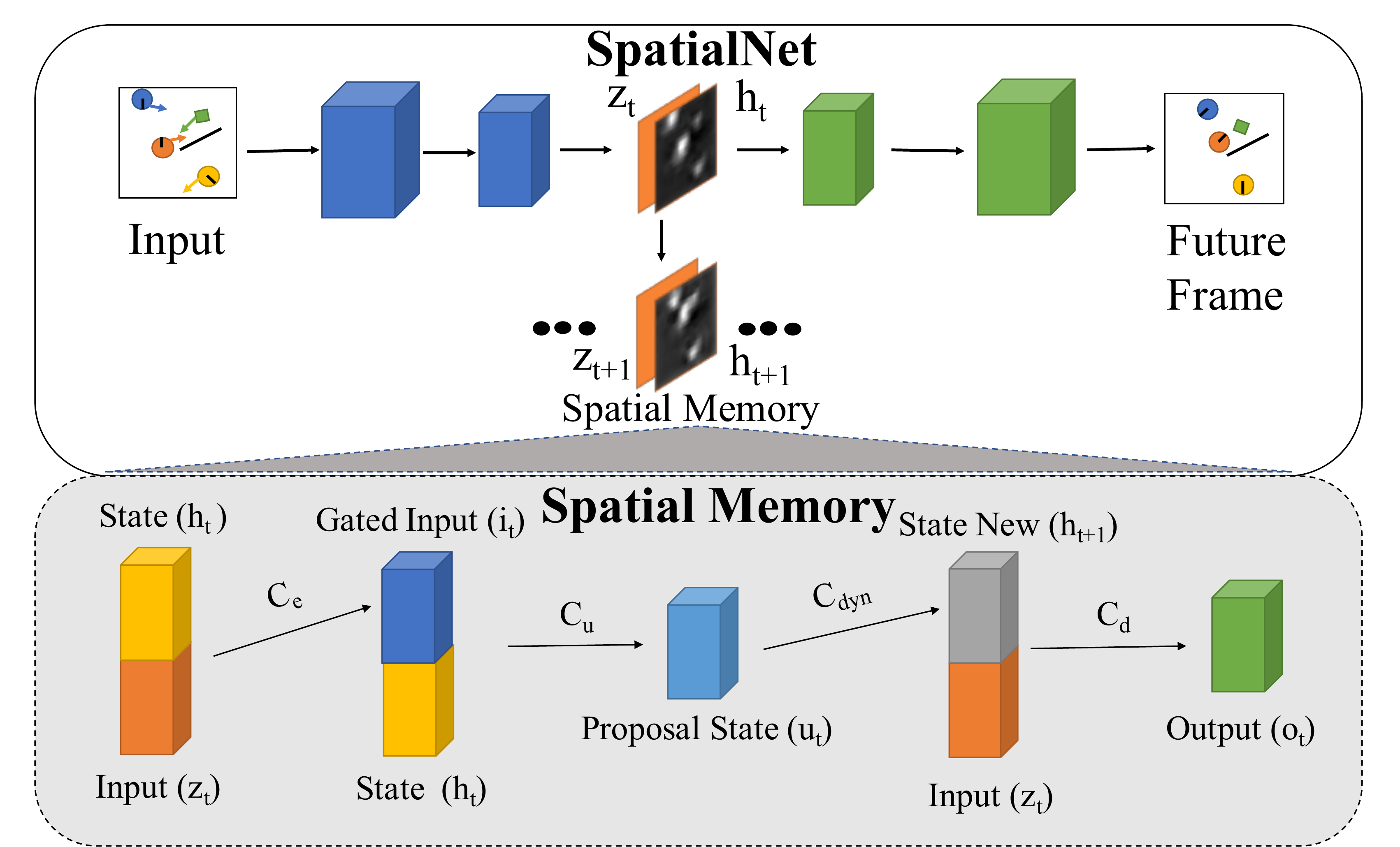}
\caption{\small Overview of the \spatial architecture. \spatial takes an RBG image as input and passes it into encoder ($\mathcal{E}$) consisting of two residual blocks to form an input encoding $z_t$. $z_t$ is processed by a \emph{spatial memory} module ($\sigma$) to obtain an output representation $o_t$, which is used by the decoder ($\mathcal{D}$) to predict the next frame. The spatial memory stores meta information about each entity and its locality. See Section~\ref{sec:spatial} for more details. 
}
\label{fig:spatial_model}
\vspace{-4mm}
\end{figure}
\textbf{\spatial Architecture} 
\spatial is conceptually simple and consists of three modules (Figure~\ref{fig:spatial_model}). The first module is a standard convolutional encoder $\mathcal{E}$ that converts an input image $x_t$ into a 3D representational map $z_t$. The second module is a spatial memory block, $\sigma$, that converts $z_t$ and the hidden state $h_{t}$ from the previous timestep into an output representation $o_t$ and new hidden state $h_{t+1}$. Finally, we have a convolutional decoder $\mathcal{D}$ that predicts the next frame $x_{t+1}$ from $o_t$. Both the encoder and decoder modules ($\mathcal{E}$ and $\mathcal{D}$) use two convolutional layers each with residual connections.


We implement the spatial memory block $\sigma$ as a 2D convolution operation. The module takes in a previous hidden state $h_{t}$ and input $z_t$ at timestep $t$, both of shape $k \times n \times n$ where $k$ is the number of channels and $n \times n$ is the dimensionality of the grid. We then perform the following operations:
\begin{equation}
\begin{split}
i_t = f(C_{e} \oplus [h_{t}; z_t]) ; ~~~ u_t = f(C_{u} \oplus [i_t; h_{t}]) \\ h_{t+1} = f(C_{dyn} \oplus u_t); ~~~ o_{t} = f(C_{d} \oplus [z_t; h_{t+1}])  
\end{split}
\label{eq:spatialnet_updates}
\end{equation}
where $\oplus$ denotes a convolution, $[;]$ denotes concatenation, $C_{e}$, $C_{u}$, $C_{dyn}$, $C_{d}$ are convolutional kernels and $f$ is a non-linearity (we use ELU~\citep{clevert2015fast}). The module first encodes a combination of $z_t$ and $h_{t}$ into a proposal state $u_t$, using two convolutions $C_{e}, C_{u}$. $C_{dyn}$ acts like a dynamics simulator and generates a new hidden state $h_{t+1}$, which captures the localized predictions for the next state around each entity. Finally, $C_d$ uses $h_{t+1}$ and $z_t$ to produce $o_t$, encoding information about the entire frame to be rendered by subsequent decoding.

Intuitively, the \spatial architecture biases the module towards storing relevant physics information about each entity in a block of pixels at the entity's corresponding location. This information is sequentially updated through the convolutions, while static information such as background texture is passed directly through the input encoding $z_t$ (see Figure 5 of appendix). We note that our spatial memory is not action-conditional, which allows us to learn from task-independent videos, as well as generalize better to new environments. Given training videos $D = \left \{(x_1^{(i)}, x_2^{(i)}, \ldots, x_{T_i}^{(i)}\right \}_{i=1}^{N}$, we learn the parameters of the model using a standard MSE-based loss function, $L(\theta) = \sum \limits_i \sum \limits_j \| \hat{x}^i_j - x^i_j \|^2 $ \label{sec:mseloss}.

\spatial is inspired by the \clstm model but is different from \clstm in that while \clstm performs an additive state updation operation ($c_t = f_t \cdot c_{t-1} + i_t \cdot \tanh(W_{cx}x_t + W_{hc}h_{t-1} + b_c)$), \spatial uses convolutions to update the hidden state (Eqn.~\ref{eq:spatialnet_updates}). This allows \spatial to better simulate moving objects and physical interactions. Another difference is that \spatial has residual connections, which provides a more straightforward inductive bias towards maintaining both static and dynamic information across states. 

\textbf{Ego-dynamics} 
One important feature of our dynamics predictor is that it is not conditioned on the action(s) of the agent, i.e. it does not account for ego-dynamics. We make this choice in order to make the dynamics prediction model task-agnostic. As we demonstrate in our experiments (Section~\ref{sec:rlexp}, this makes our approach generalize well to a variety of different tasks, and learn faster in transfer experiments. 

\section{Experiments}
\label{sec:exp}
We perform two empirical studies to evaluate our hypothesis. First, we evaluate various frame prediction models, including our proposed \spatial, in terms of their capacity to predict future states and model physical interactions (Sections~\ref{sec:physexp} and ~\ref{sec:physexp2}). Then, we investigate the use of these dynamics predictors for policy learning in different environments (Section~\ref{sec:rlexp}). 

\paragraph{Physics video dataset}
\label{sec:physdata}
In order to train a prediction model specifically for physical interaction, we generate a new video dataset, \emph{PhysVideos}, using a 2-D physics engine~\cite{pymunk}. Each video in the dataset has frames of size $84 \times 84 \times 3$ with 4-8 different shapes (such as squares or circles) moving inside a room with  up to 3 randomly generated interior walls (see Figure \ref{fig:teaser} (top)). Objects are initialized with random positions and velocities, a friction coefficient of 0.9 and elasticity of 0.95, resulting in diverse object movements across each  trajectory. Being able to predict the future in this type of environment requires 2-dimensional physics reasoning, such as inferring velocity from past movement, anticipating changes in momentum due to collisions, and predicting rotations of each object. We generate 5000 different trajectories in total -- 4500 for training a dynamics predictor and 500 for testing -- with each trajectory having a length of 125 steps. See supplementary material for sample trajectories.


\subsection{Frame Prediction}
\label{sec:physexp}
In this section, we evaluate various frame prediction models on their accuracy across different horizons. We report results on the 500 trajectories from the test set of \emph{PhysVideos}.

\textbf{Baselines}
We compare our model, \spatial, with the following baselines: 
\begin{enumerate}
    \item \textit{\ohnet}: the model of \cite{Oh2015Action} modified to work without action-conditioning, i.e. $h_t^{dec} = h_t^{enc}$. 
    \item \textit{\clstm}~\cite{xingjian2015convolutional}: this model replaces all the inner operations of an LSTM with convolutions. We use a kernel size of 5 and the same encoders and decoders as in \spatial.
    \item \textit{\clstm + Residual}: a modified version of \clstm with added residual connections from input to output of the LSTM cell.
\end{enumerate}
We train all prediction models using mean squared error (MSE) loss. We use the Adam optimizer~\citep{Kingma2015Adam} in our experiments with a learning rate of $10^{-4}$.

\begin{table}[t]
\centering
\resizebox{\columnwidth}{!}{%
\begin{tabular}{lccccc}
    \toprule
    \textbf{Model} & \textbf{1 step} & \textbf{5 step} & \textbf{10 step} & \textbf{Objects Lost} \\
    \midrule
    \textit{\ohnet~\citep{Oh2015Action}}  & 0.0061  & 0.0140 & 0.0268 & 1.0  \\
    \textit{\clstm~\citep{xingjian2015convolutional}} & 0.0026  & 0.0303 & 0.0503 & 0.4\\
    \textit{\clstm + Residual} & 0.0026 & 0.0141 & 0.0210 & 0.45 \\
    \textit{\spatial} & \textbf{0.0024} & \textbf{0.0114} &\textbf{ 0.0176} & \textbf{0.13} \\
    \bottomrule
\end{tabular}
}
\caption{\small MSE for multi-step prediction on PhysVideos (lower is better). All models were trained with 1 step prediction loss. \spatial suffers least from compound errors during prediction, and is able to maintain objects and dynamics more consistently (Figure~\ref{fig:physdata_qual}). Number of objects lost (after 20 steps) was determined manually by evaluating 15 random videos in the test set.}
\label{tbl:phys_pred}
\end{table}
\begin{figure}[t]
\centering
\includegraphics[width=\linewidth]{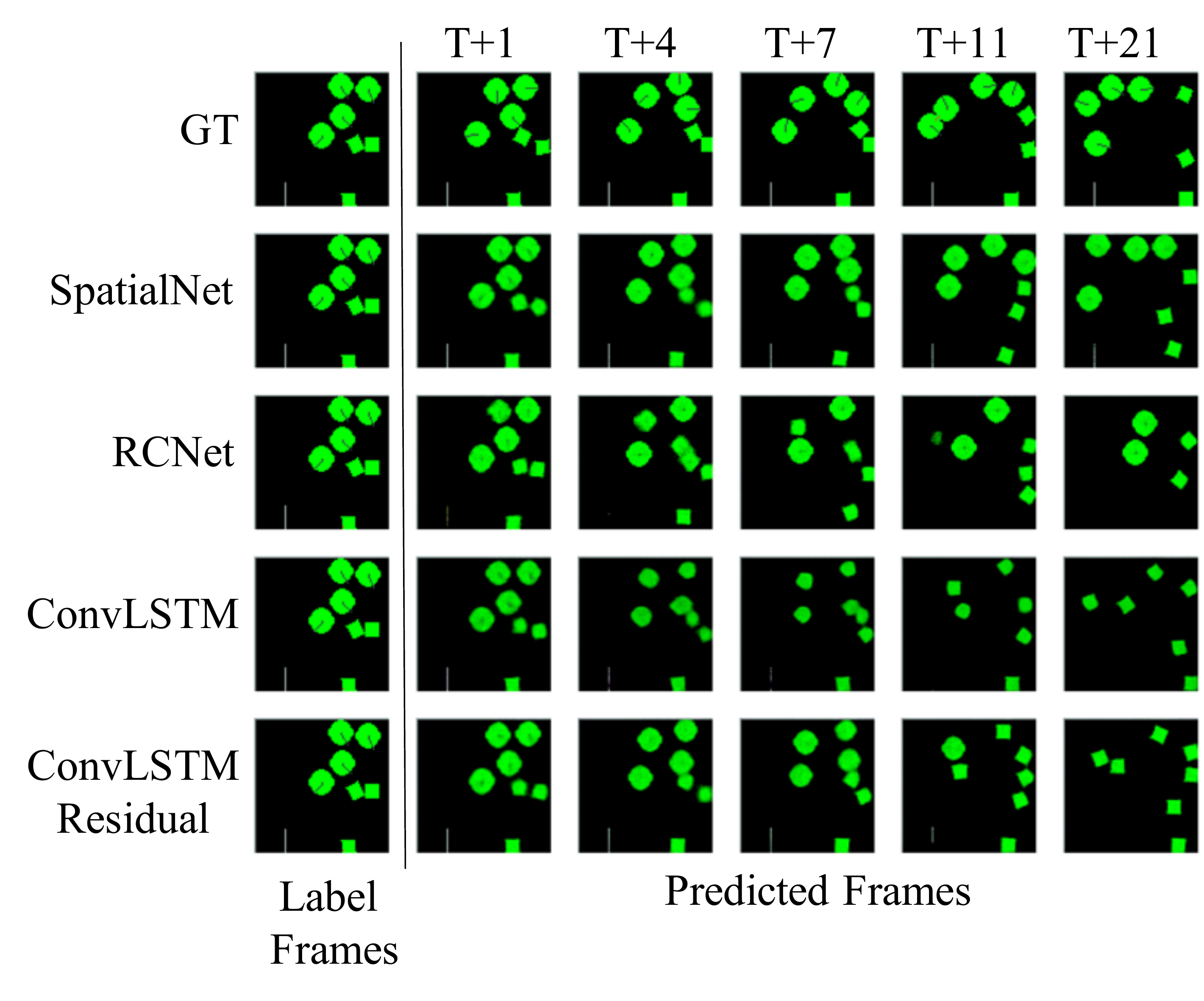}
\caption{\small Visualization of multi-step predictions of \spatial, \ohnet, and \clstm variants, along with ground truth (GT). After 20 steps of self prediction, \spatial maintains the internal wall and all seven objects in the scene while \ohnet~\citep{Oh2015Action} loses the internal wall and 3 of the moving objects. \clstm loses shape information and has less accurate dynamics prediction. \spatial is most consistent in obeying physical laws.}
\label{fig:physdata_qual}
\end{figure}

\textbf{Results} 
From Table~\ref{tbl:phys_pred}, we see that \spatial achieves a lower test MSE compared to all the baselines, especially for multi-step predictions. This suggests that \spatial encourages better dynamic generalization compared to \ohnet and \clstm. We can also observe from Figure \ref{fig:physdata_qual}, that \spatial is able to accurately maintain the number of objects in the video even after 20 steps, while the baselines suffer from merging of objects (\ohnet) or loss of shape information (\clstm). Further, \spatial is also able to maintain background details such as walls that are quickly lost in \ohnet, as the spatial memory structure allows the input to easily remember fixed background information. We also find that the spatial memory's overall structure allows it to be very resistant to input noise as well as better generalize to unseen environments -- please see the supplementary material for detailed analyses.


\subsection{Predicting physical parameters}
\label{sec:physexp2}
To further probe the representations learned by the frame prediction models, we test their ability to predict physical properties of environments (e.g. elasticity or drag) from videos. We train a 2 layer classification model on top of the hidden state representations produced by \spatial/\clstm to predict one of 3 values for elasticity/drag -— low, medium or high. Only the classification layers are trained, while the rest of the parameters are kept fixed (except for \textit{full train}). 

From Table~\ref{tbl:phys_finetune}, we see that randomly initialized parameters or \spatial trained on Atari Pong don't do well, indicating that they don't capture physics. \spatial trained on PhysVideos gets an accuracy of around 69\% on drag prediction (close to the fully trained model accuracy of 78\%). This shows that the pre-training indeed helps the model acquire priors over physical dynamics. Further, the low numbers of the model trained on Atari Pong indicate that task-specific frame prediction may not generalize well.

\begin{table}[t]
\setlength{\tabcolsep}{5.5pt}
\centering
\begin{tabular}{ccc}
    \textbf{Model} & \textbf{Drag} & \textbf{Elasticity}\\
    \midrule
    \textit{\spatial (random init)} & 35.8 &  43.8\\
    \textit{\spatial (PT on Atari Pong)} & 35.0 & 33.6 \\
    \textit{\clstm (PT on PhysVideos)} & 57.2 & 53.2 \\
    \textit{\spatial (PT on PhysVideos)} & 69.8 & 56.9 \\
    \midrule
    \textit{\spatial (full train)} & 78.5 & 67.8\\

    \bottomrule
\end{tabular}
\caption{Accuracies on predicting drag and elasticity from video frames (PT = pre-training)}
\label{tbl:phys_finetune}
\vspace{-10pt}
\end{table}
\subsection{Reinforcement Learning}
\label{sec:rlexp}
In this section, we describe the use of \spatial to accelerate reinforcement learning. We first train \spatial on the physics video dataset described in the previous section. Then, we use the pre-trained \spatial model as a future frame predictor for a reinforcement learning agent. We perform empirical evaluations on three different platforms - a suite of 2D games (PhysWorld), a 3D partially observable environment, and a stochastic version of Atari games~\citep{machado2017ale}. We demonstrate that \agent with \spatial pre-training outperforms existing approaches in all platforms. The \agent architecture also allows for effective decoupling of environment dynamics from agent policy, which results in better transfer performance across tasks.




\myparagraph{Experimental setup}
In our experiments, we use \spatial to predict the next k\footnote{We find k=3 to work well in our experiments.} future frames. We then stack the current frame with the k predicted frames and use this as input to a model free policy.  We use the Adam optimizer with learning rate $10^{-4}$ to train model predictions and the same set of hyper-parameters for training all policy agents as those used in \cite{Schulman2017Proximal}. For our policy network, we use the architecture described in \cite{Mnih2015Human}. We report numbers averaged over 3 different random seeds. 

\myparagraph{Baselines}
We compare our agent (\agent) with a number of different baselines:
\begin{enumerate}
     \item \textit{PPO}: A standard implementation of Proximal Policy Gradient (PPO)~\citep{Schulman2017Proximal}, which is model-free and uses the current frame with the last k frames to output an action. The number of frames provided to PPO is the same as that provided to \agent.
     \item \textit{PPO + VF}: PPO with value function expansion~\cite{feinberg2018model}, which uses a dynamics predictor to obtain a more consistent estimate of the current state's value.
    \item \textit{I2A}: Imagination Augmented Agent~\cite{weber2017imagination} uses a combination of past frames and a recurrent encoding of future rollouts\footnote{Rollouts are $k$ future frames predicted by \spatial.} as input to the policy. 
    \item \textit{ISP}: A variant of \agent that uses the hidden layer of \spatial directly as input to a policy network. 
    \item \textit{JISP} : ISP with auxiliary frame prediction loss. 
    \item \textit{Other frame predictors}: Finally, we also consider baselines where we augment our agent, \agent, with future frames predicted by \ohnet~\citep{Oh2015Action} and \clstm~\citep{xingjian2015convolutional}. 
\end{enumerate}


\myparagraph{PhysWorld}
\begin{table}[t]
\small
\setlength{\tabcolsep}{5.5pt}

\centering
\resizebox{0.5\textwidth}{!}{

\begin{tabular}{lccc}
      & \textit{PhysGoal} &  \textit{PhysForage} &  \textit{PhysShooter} \\
    \midrule
    PPO & 17.9 (0.8) & 44.2 (5.4) & 23.2 (1.2) \\
    PPO + VF & 19.2 (2.4) & 40.4 (5.4) & 26.1 (2.9) \\ \midrule
    I2A + \spatial (action-cond) & 4.2 (0.4) & 23.7 (3.1) & 16.5 (1.8)\\
    I2A + \spatial &  16.4 (6.2) & 20.8 (2.0) & 19.3 (0.7) \\  \midrule
    \agent + \ohnet & 20.7 (3.1) & 46.3 (23.4) & 31.7 (1.0)\\
    \agent + \clstm & 21.6 (2.1) & 39.5 (7.0)  &  29.1 (1.6)\\ \midrule
    ISP & 15.2 (1.2) & 45.3 (5.5) & 18.6 (1.1)\\
    JISP & 18.2 (5.5) &  \textbf{124.3} (27.1) & 28.6 (1.5) \\ \midrule
    \agent + \spatial (Blink) &  24.6 (2.8) & 48.5 (5.3) & 31.0 (1.9) \\
    \agent + \spatial (PhysVideos) & \textbf{30.8} (5.2) & 50.6 (11.5) &  \textbf{42.3} (2.9)\\
    \bottomrule
\end{tabular}
}
\caption{\small Average scores (with standard deviation) obtained in PhysWorld environments by various agents after 10 million frames of training. Scores are rewards over 100 episodes, averaged over runs with 3 different random seeds. \agent + \spatial consistently outperforms the other approaches. \ohnet, \spatial, \clstm are pretrained on PhysVideos. PPO+VF = PPO with Value Function Expansion. \spatial(Blink) refers to a model trained on videos with blinking objects. We add 500K additional frames to the PPO baselines to account for the  frames used in pre-training for the other models.}
\label{tbl:physworld_quantative}
\vspace{-3mm}
\end{table}

We first consider PhysWorld, a new collection of three physics-centric 2D games that we created. These games require an agent to accurately predict object movements and rotations in order to perform well. All three tasks have an environment consisting of around 10 randomly moving boxes and circles as well as up to three internal impassable walls. \emph{PhysGoal} is a navigation task to reach goals while avoiding objects, \emph{PhysForage} is an object gathering task, and \emph{PhysShooter} requires a stationary agent to shoot selected moving objects while preventing collisions.  The objects in each of these environments are \emph{different colors and sizes} than those used to train the dynamics predictor in Section~\ref{sec:physdata}. We provide a detailed description of each task in the supplementary material. We emphasize that the main parameters (like object velocities, rotations,etc.) in the PhysWorld games are fully \textbf{randomized} for each episode. To obtain good performance, agents need a good understanding of general physics and cannot just memorize frames.

\begin{figure}
\centering
\begin{minipage}[t]{0.5\columnwidth}
\includegraphics[width=1.0\linewidth]{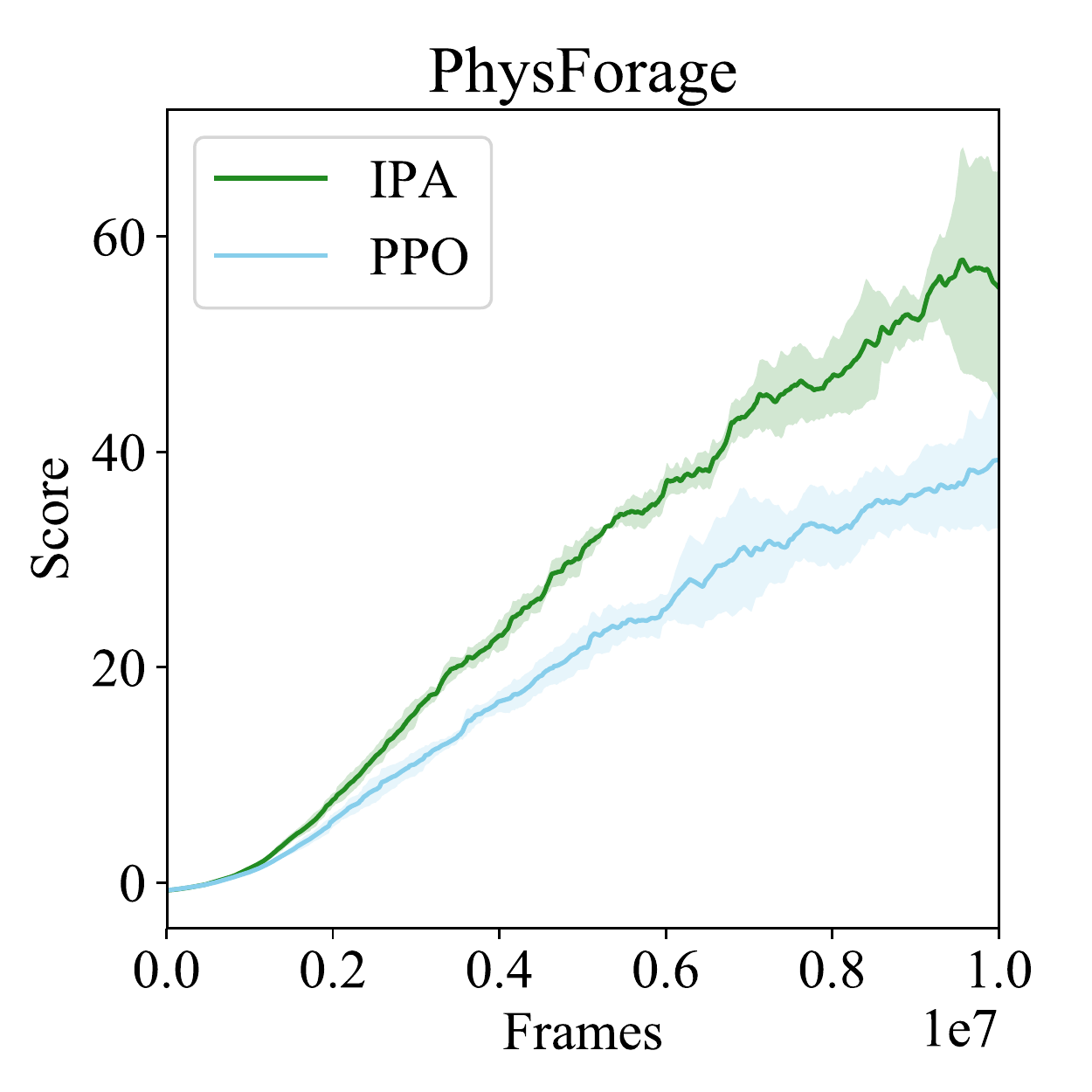}
\end{minipage}%
\begin{minipage}[t]{0.5\columnwidth}
\includegraphics[width=1.0\linewidth]{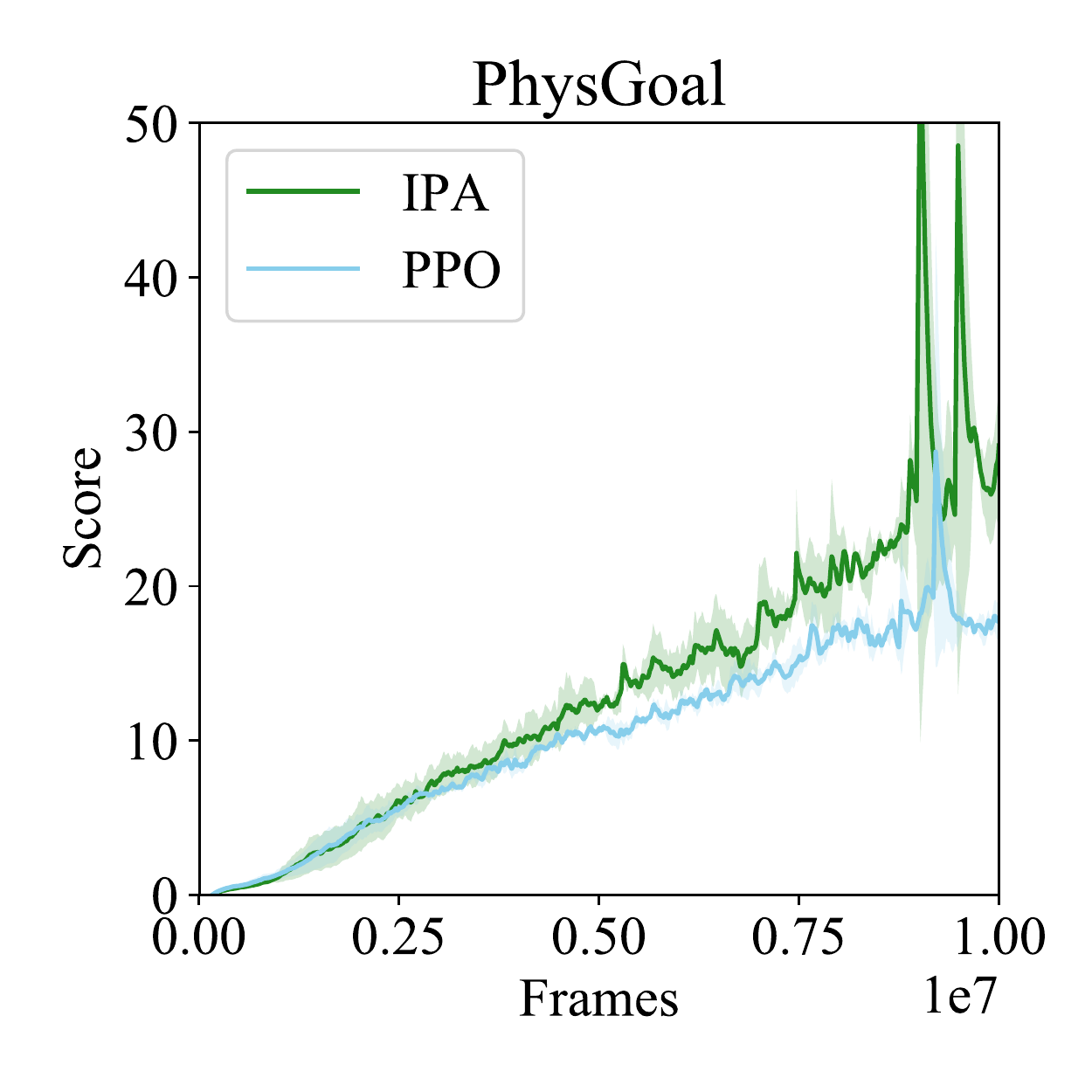}
\end{minipage}

\begin{minipage}[t]{0.5\columnwidth}
\includegraphics[width=1.0\linewidth]{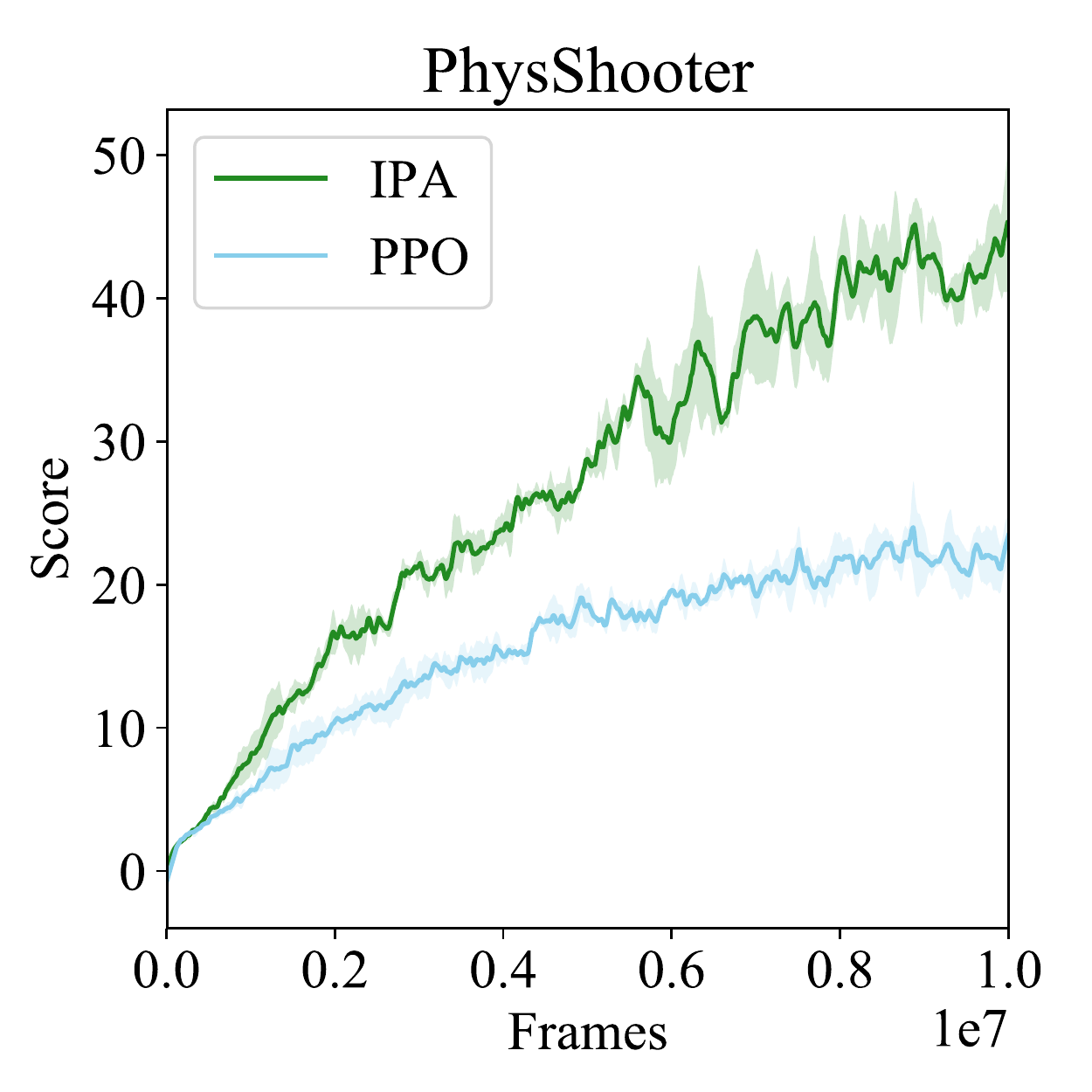}
\end{minipage}%
\begin{minipage}[t]{0.5\columnwidth}
\includegraphics[width=1.0\linewidth]{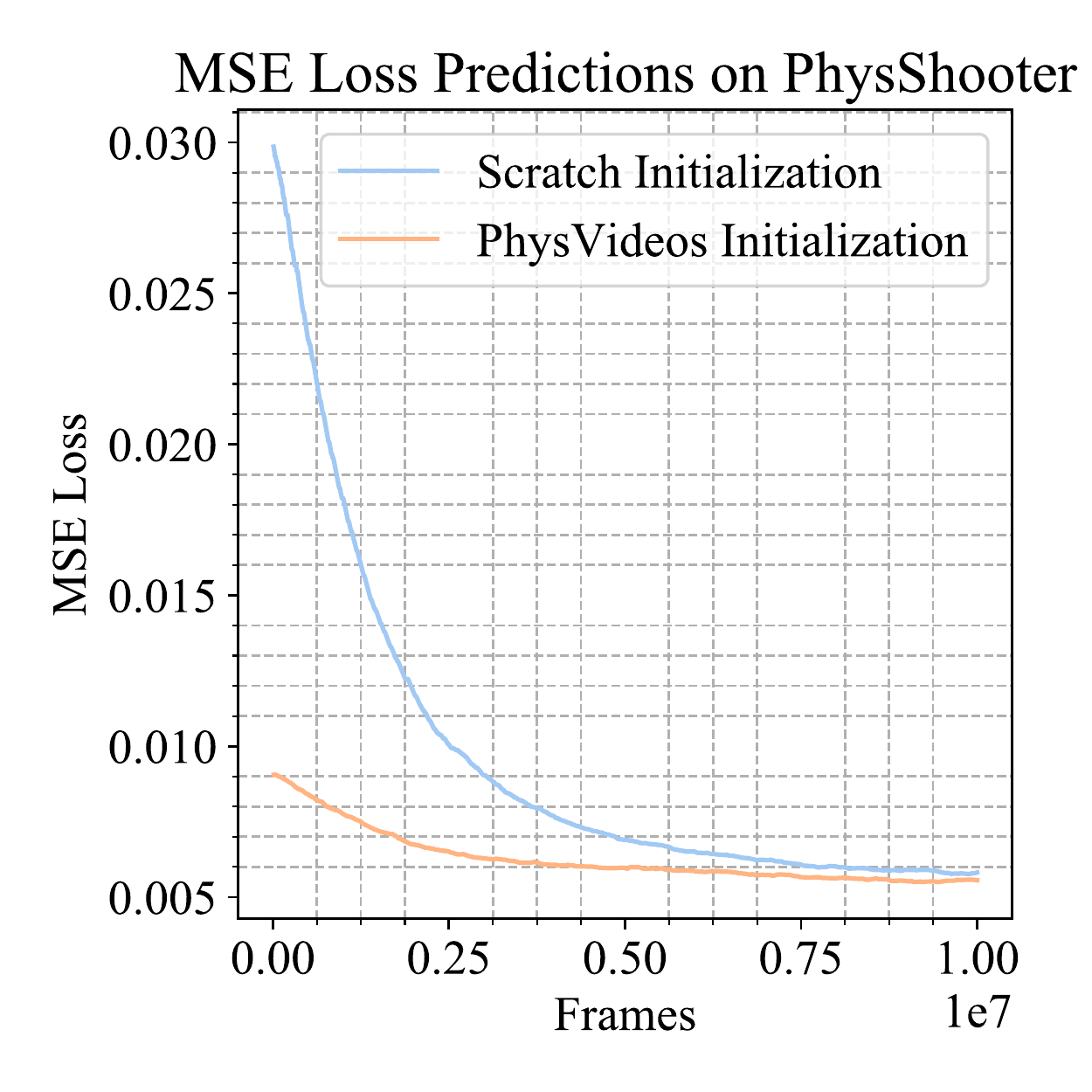}
\end{minipage}

\caption{Training curves on PhysWorld and MSE curve (bottom-right) for predicting future frames in PhysShooter.}
\label{fig:phys_plot}
\vspace{-10pt}

\end{figure}
\textit{Results:}
We detail the performance of our approach compared to the baselines in Table~\ref{tbl:physworld_quantative} and show learning curves in Figure~\ref{fig:phys_plot}. Quantitatively, we find that our approach, IPA + SpatialNet (PhysVideos), obtains significant gains over most baselines in all three tasks in PhysWorld using \agent with SpatialNet. We find that \agent with \ohnet or \clstm provides less benefits, due to slower learning than SpatialNet. We also find PPO with value expansions (PPO+VF) also provides slight gains, but significantly less than the gains conferred by \agent. I2A leads to no gains in performance, since it generates a global encoding of an image, destroying local dynamics information of objects. Both ISP and JISP perform worse than \agent except on PhysForage. On PhysForage, we find that JISP performs better, likely due to increased policy capacity compared to \agent (i.e. more parameters). We observe that \spatial trained on videos with blinking objects does not provide as much of a benefit, pointing to the fact that our full model is learning some aspects of dynamics beyond just object appearances. 

\agent encourages the policy to take into account the future physics of objects, a bias crucial for good performance on each of the PhysWorld environments. Qualitatively, we observe that in all three environments, \agent agents navigate to goals and collect objects with  more confidence, even if there are nearby obstacles nearby. In PhysShooter, \agent agents are much more able to hit objects further away on the map, which require multiple time-steps before collisions. Figure~\ref{fig:phys_plot} demonstrates how having a good prior results in faster learning of the environment dynamics of PhysShooter. 

\label{sec:physworld_results} 

Figure~\ref{fig:phys_plot} shows the relative training rates of policies under PPO and \agent. In \textbf{Phys-Shooter} we see immediate benefits in using a physics model, as physics knowledge of the future is crucial as the agent only gets one action approximately every 4-5 frames. In \textbf{Phys-Goal} and \textbf{Phys-Forage}, we see long term benefits in knowing future physics as this knowledge allows the agents to more efficiently collect points.

\myparagraph{PhysShooter3D}
Additionally, we also evaluate on PhysShooter3D, a 3D physics game which we construct using Bullet~\citep{Coumans2010Bullet}. We add gravity to the world and generate moving projectiles that follow bouncing parabolic trajectories. We then render 2D images from a particular viewpoint, causing moving objects to be partially or fully occluded at times. With these additional factors, learning dynamics is even more challenging. The game requires a stationary agent to fire bullets at selected 3D projectiles without itself being hit by any projectiles. We found that PPO obtained a score of $0.86 \pm 0.28$ while IPA + \spatial obtained $1.73 \pm 0.09$ and IPA using Ground truth frames obtained $4.16 \pm 0.84$. This demonstrates that IPA generalizes well to partially observed settings, with still room for improvement by performing better frame prediction.

\myparagraph{Stochastic Atari Games}
In addition to PhysWorld and PhysWorld3D, we also investigate the performance of \agent on a stochastic version of the Arcade Learning Environment (ALE)~\citep{bellemare2013arcade}, by adding \emph{sticky actions}, where an agent repeats its last action with probability $p=0.5$. This stochasticity was shown to be the most challenging type of randomization to add to ALE~\citep{hausknecht2015impact,machado2017ale}. We evaluate performance on all Atari games, a subset of which are shown in Table~\ref{tbl:atari_quantative}. All Atari experiments are run with 5 different seeds. 

We emphasize that this is an \textit{out-of-domain} evaluation -- we use the prior trained on PhysVideos to initialize the dynamics predictor for Atari, which contains a significantly different pixel space.  Further, not all Atari games are reliant on understanding physics and we do not expect our approach to provide significant gains on those environments. 

\textit{Results:}
From Table~\ref{tbl:atari_quantative}, we observe that \agent outperforms PPO in 8 out of the 10 different tasks\footnote{Results on all Atari games are in supplementary material.} -- these are all games that contain physical interactions between objects and benefit from our prior. In several games like Enduro, Breakout, Frostbite, FishingDerby and Assault, \agent provides benefits later on in training after the agent has figured out a good initial policy. In others like Asteroids and DemonAttack, \agent shows immediate boosts in training performance, resulting in faster policy learning. On Pong, where \agent performed worse than PPO, we found that the agents learned to place paddles at one particular location where without paddle movement, the ball would bounce and score points. Similarly, on Ice Hockey, where PPO outperformed IPA, we found that agents can learn a repetitive strategy to prolong the game indefinitely, removing the need for tracking dynamics information. Under such situations, there is no added advantages to predicting dynamics, explaining the reduced scores of \agent. We provide additional qualitative results, including frame predictions, in the supplementary material.

\begin{table}
\small
\centering
\resizebox{\columnwidth}{!}{
\begin{tabular}{lcccc}
 & PPO & I2A & IPA \\
\midrule
Assault & 2932 (153) & \textbf{3249.7} (378) & 2968.4 (124) \\
Asteroids & 1321 (233.5) & 1340 (351) & \textbf{2098} (102) \\
Breakout & 19.7 (0.9) & 18.7 (0.0) & \textbf{23.4} (1.0) \\
DemonAttack & 5510 (412) & 5492 (233) & \textbf{6793} (558)  \\
Enduro & 376.7 (10.5) & 380 (8.0) & \textbf{398.6} (23.0)  \\
FishingDerby & 6.7 (10.1) & \textbf{12.1} (4.0) & 9.3 (3.0) \\
Frostbite & 1342 (2154) & 1649 (2100) & 1701 (2485)  \\
IceHockey & -5.9 (0.3) & -6.3 (0.0) & -6.1 (0.0)  \\
Pong & \textbf{6.6} (14.1) & -1.4 (15.0) & 2.2 (13.0)  \\
Tennis & -6.3 (2.1) & -8 (4.0) & \textbf{-3.8} (1.0) \\
\bottomrule
\end{tabular}

}
\caption{\small Scores (and standard deviation) obtained on Stochastic Atari Environments with \emph{sticky actions} (actions repeated with 50\% probability at each step). Scores are average performance over 100 episodes after 10M training frames, over 5 different random seeds with included standard deviations.
}
\label{tbl:atari_quantative}
\end{table}

\begin{table}[h]
\small
\setlength{\tabcolsep}{5.5pt}

\centering
\resizebox{\columnwidth}{!}{

\begin{tabular}{lcccc}
     Source env & Agent & Model  & Policy  & Reward\\
      &  &  transfer &  transfer & \\
     \midrule
    \textit{None} & PPO & - & - & 23.2 \\
     \textit{None} & \agent & - & - & 35.42  \\
     \textit{PhysVideos} & \agent + \spatial & Y & - &  42.27\\
    \midrule
     \multirow{4}{*}{PhysGoal} & PPO & - & Y & 25.42 \\
      & IPA + \spatial (Fix) & Y & N & 26.30 \\
     &  IPA + \spatial (FT) & Y & N & \textbf{42.83}\\
     &  IPA + \spatial (FT) & Y & Y & 42.44  \\
    \midrule
    \multirow{4}{*}{PhysForage}  & PPO & - & Y & 24.47 \\
     &  IPA + \spatial (Fix) & Y & N & 30.30\\
     & IPA +  \spatial (FT) & Y & N & \textbf{53.66} \\
     & IPA +  \spatial (FT) & Y & Y & 40.40 \\

    \bottomrule
\end{tabular}


}
\caption{\small Effects of model initialization and transfer on training policies in \textit{PhysShooter}.
Topmost section shows baseline PPO, random initialization of dynamics for IPA, and pre-trained IPA using PhysVideos. The bottom two sections demonstrate results while transferring different models from two other games -- direct policy (PPO), transfer dynamics model and fix it (Fix), transfer dynamics and finetune (FT), and transfer both dynamics+policy and finetune.
\agent allows decoupling of policy transfer from model transfer, allowing better transfer in cases of environment similarity but task dissimilarity.  Scores obtained on the PhysWorld environments after training for 10M frames and evaluated by taking average rewards of the last 100 training episodes.
}
\label{tbl:physworld_transfer}
\vspace{-15pt}
\end{table}

\subsection{Transfer and Generalization}
\label{sec:physworld_transfer}
We now present some empirical results under the transfer scenario and provide some analysis of our model. Table~\ref{tbl:physworld_transfer} also shows the impact of initializing \agent with different pre-trained dynamics models on the PhysShooter environment. We find that initializing \spatial with random parameters does not perform very well, but using a \spatial pretrained on PhysVideos provides better performance (see Figure~\ref{fig:phys_plot} for MSE errors). Moreover, we observe that transferring a \spatial model fine-tuned on a different task like PhysForage/PhysGoal results in even greater performance improvements.
\emph{Interestingly, we note that transferring just the dynamics model in \agent results in a larger performance gains than transferring both model and policy}. For instance, transferring the model from PhysForage results in a score of $53.7$ while transferring both model+policy gets a lower score of $40.4$. The former is a 27\% increase compared to using just PhysVideos ($42.27$) , while the latter results in a lower score. This provides further evidence that decoupling model learning from policy learning allows for better generalization.

\section{Conclusion}
We have proposed a new approach to model-based reinforcement learning by learning task-agnostic dynamics priors. First, we pre-train a frame prediction model (\spatial) on raw videos of a variety of objects in motion. We then use this network to initialize a dynamics model for an RL agent, which makes use of predicted frames as additional context for its policy. Through several experiments on three different domains, we demonstrate that our approach outperforms model-free techniques and approaches that learn environment dynamics from scratch. We also demonstrate the generalizability of our dynamics predictor through transfer learning experiments. 


\paragraph{Acknowledgements}
We would like to thank Alexander Botev, John Schulman, Tejas Kulkarni, Bowen Baker and the OpenAI team for providing helpful comments and suggestions.


{\small
\bibliographystyle{plainnat}
\bibliography{reference,phys_reinforce}
}



\appendix
\renewcommand{\thesection}{A.\arabic{section}}
\renewcommand{\thefigure}{A\arabic{figure}}
\setcounter{section}{0}
\setcounter{figure}{0}

\section{Additional Dynamic Prediction Experiments}

\begin{table}[h]
\setlength{\tabcolsep}{5.5pt}
\centering
\begin{tabular}{lccc}
    \toprule
     $\epsilon$ & \ohnet & \clstm & \spatial (ours) \\
    \midrule
     0 & 0.0061 & 0.0026 &\textbf{0.0024} \\ 
    0.1 & 0.0078 & 0.0030 & \textbf{0.0026} \\
    0.5 & 0.0268 & 0.0072 & \textbf{0.0062} \\

    \bottomrule
\end{tabular}
\caption{\small MSE loss on physics prediction data-set on on single-step prediction with test inputs corrupted with Gaussian noise of magnitude $\epsilon$ (model trained with no corruption). Due to its local nature, \spatial suffers less form errors in inputs and is able to maintain object numbers/dynamics more consistently even with domain shift.}
\label{tbl:phys_noise}
\end{table}


\subsection{Sensitivity to Corruption of Inputs}
We investigate the effects of noisy observations in the input domain at test time on both \spatial and \ohnet, by adding different amounts of Gaussian random noise to input images (Table \ref{tbl:phys_noise}).  We find that \spatial is more resistant to noise addition. \spatial predictions are primarily local, preventing compounding of error from corrupted pixels elsewhere in the image whereas \ohnet compresses all pixels into a latent space, where small errors can easily escalate.

\subsection{Qualitative visualizations of Generalization Predictions}
\begin{figure*}[h]
\centering
\includegraphics[width=0.5\linewidth]{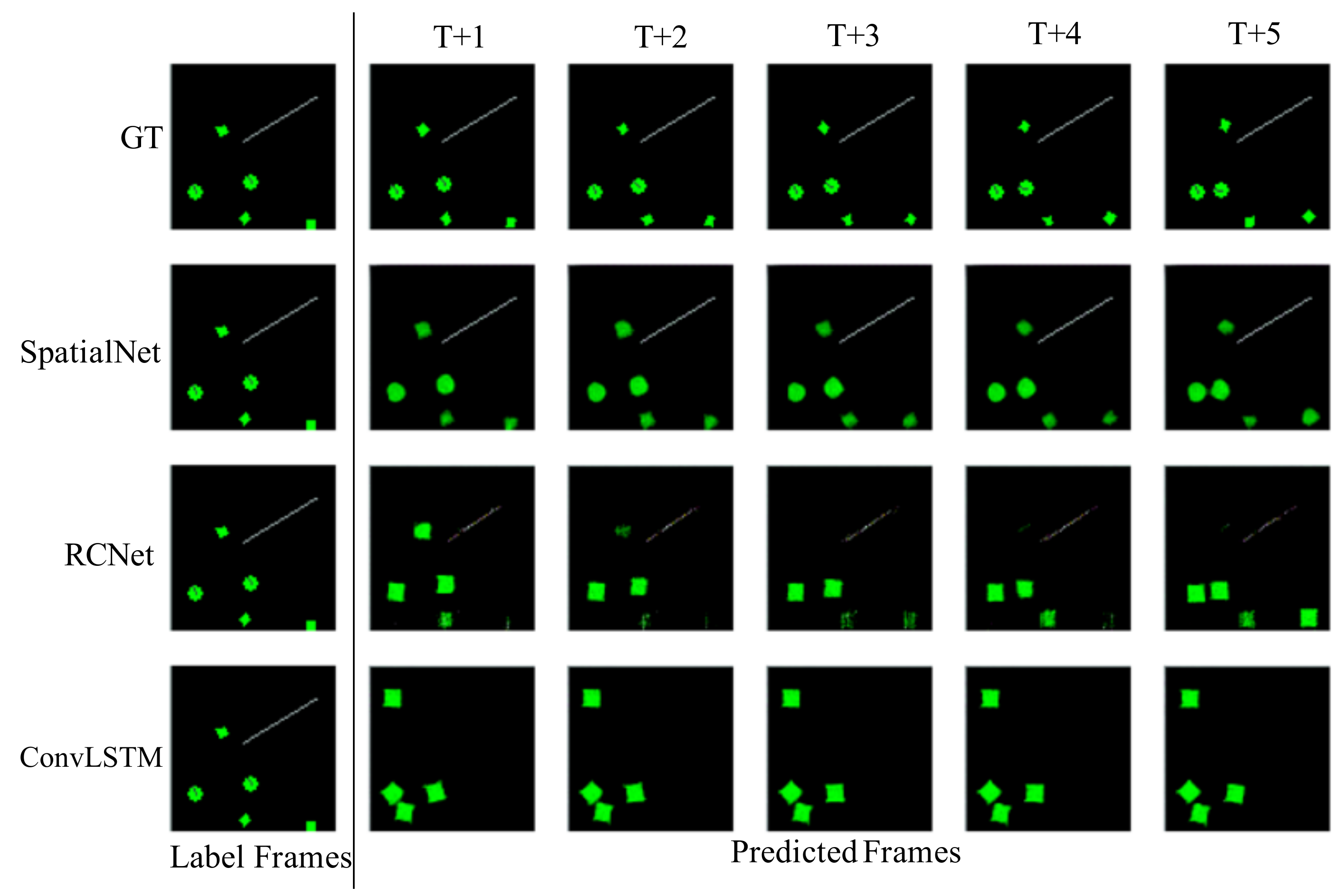}
\caption{\small Predictions of \spatial, \ohnet on test data-set with objects twice as small and with twice the movement speed as trained on. All shown frames are one step predictions. \spatial is able to accurately generalize to smaller, faster objects while \ohnet is unable to generate the shapes of the smaller objects and suffers from background degradation and \clstm is unable to maintain shapes and dynamics.} 
\label{fig:phys_gen}
\end{figure*}

\begin{figure}[t]
\centering
\includegraphics[width=\linewidth]{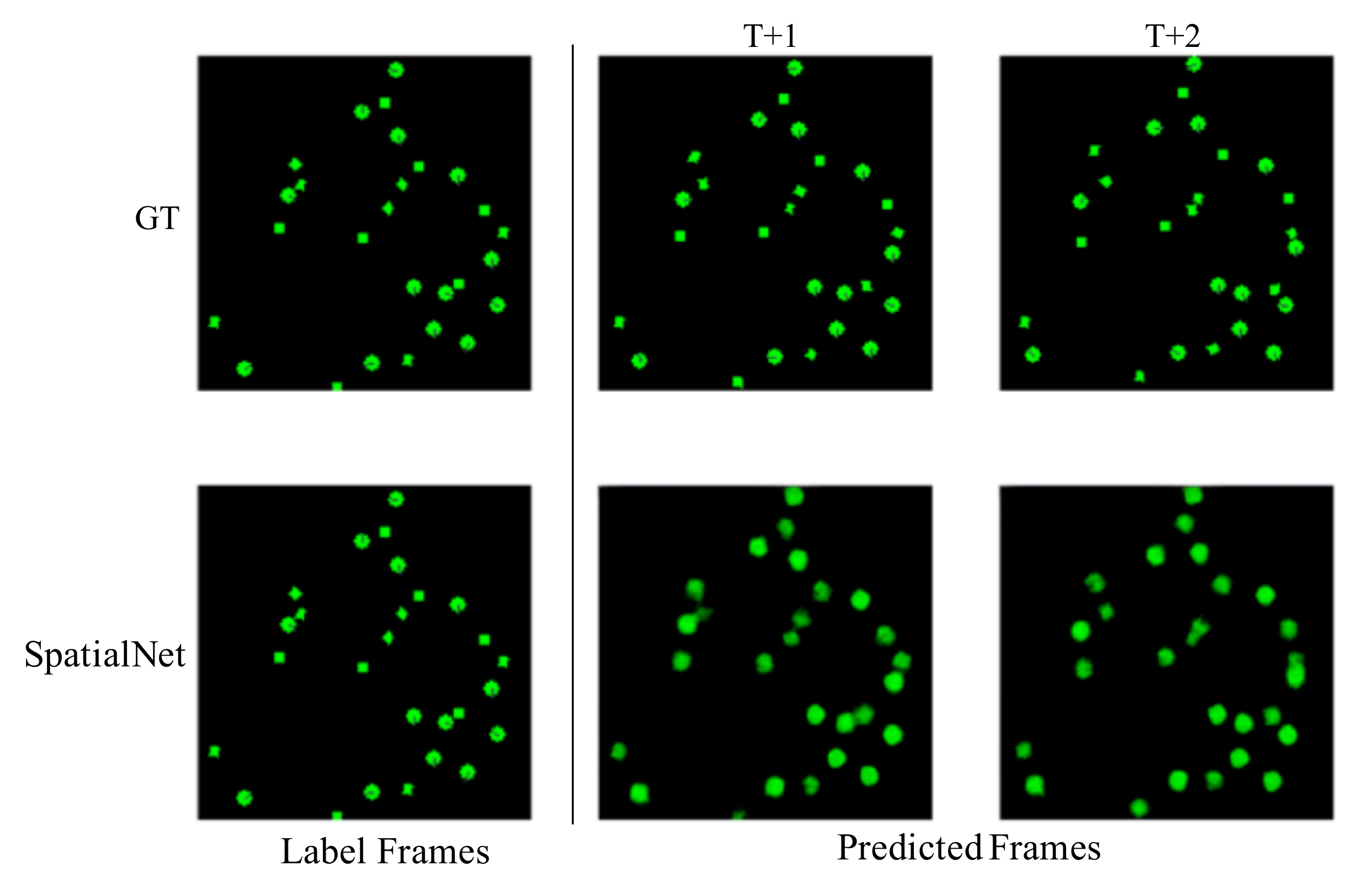}
\caption{\small Predictions of \spatial on input images of 168 x 168 when \spatial was trained on 84 x 84 images. Prediction shown are 1 step future predictions. \spatial is able to maintain physical consistency in at large input sizes.} 
\label{fig:phys_gen_large}
\end{figure}

We provide visualizations of video prediction on each of the generalization datasets in Figure~\ref{fig:phys_gen} and Figure~\ref{fig:phys_gen_large}.


\subsection{Dataset Generalization.} We test generalization by evaluating on two unseen datasets. For the first, we create a test set where objects are half the size of the training set and initialized randomly with approximately twice the starting velocity. In this new regime, we found that \ohnet had a MSE of 0.0115, \clstm has a MSE of 0.0067, while \spatial had a MSE of 0.0039. We find \ohnet is unable to maintain shapes of the smaller objects, sometimes omitting them, while \clstm maintains shape but is unable to adapt to new dynamics as seen in \fig{fig:phys_gen}. In contrast, \spatial local structure allows it to generate new shapes, and its dynamic seperation allows better generalization. In the second dataset, we explore input size invariance. We create a second testing data-set consisting 16-32 random circles and squares and input images of size 168x168x3 (the density of objects per area is conserved). On this dataset, we obtained a MSE of 0.0042 compared to \clstm of 0.0060, which is comparable to the MSE on the original test dataset of 0.0024, with qualitative images in \fig{fig:phys_gen_large} showing that the spatial memories local structure allows to easily generalize to different input image sizes.

\section{PhysWorld}
\label{sec:physworld}
We provide a description of the three games environments in PhysWorld:

\emph{PhysGoal:} In this environment, an agent has to navigate to a large red goal. Each successful navigation (+1 reward) respawns the red goal at a random location while collision with balls or boxes terminates the episode (-1 reward).

\emph{PhysForage:} Here, an agent has to collect moving balls while avoiding moving boxes. Each collected ball (+1 reward) will randomly respawn at a new location with a new velocity. Collision with boxes lead to termination of episode (-1 reward).

\emph{PhysShooter:} In PhysShooter, the agent is stationary and has to choose an angle to shoot bullets. Each bullet travels through the environment until it hits a square (+1 reward) or circle (-1 reward) or leaves the screen. If a moving ball or box hits the agent (-1 reward), the episode is terminated. After firing a bullet, the agent cannot fire again until the bullet disappears. 

\begin{figure}
\centering
\includegraphics[width=\linewidth]{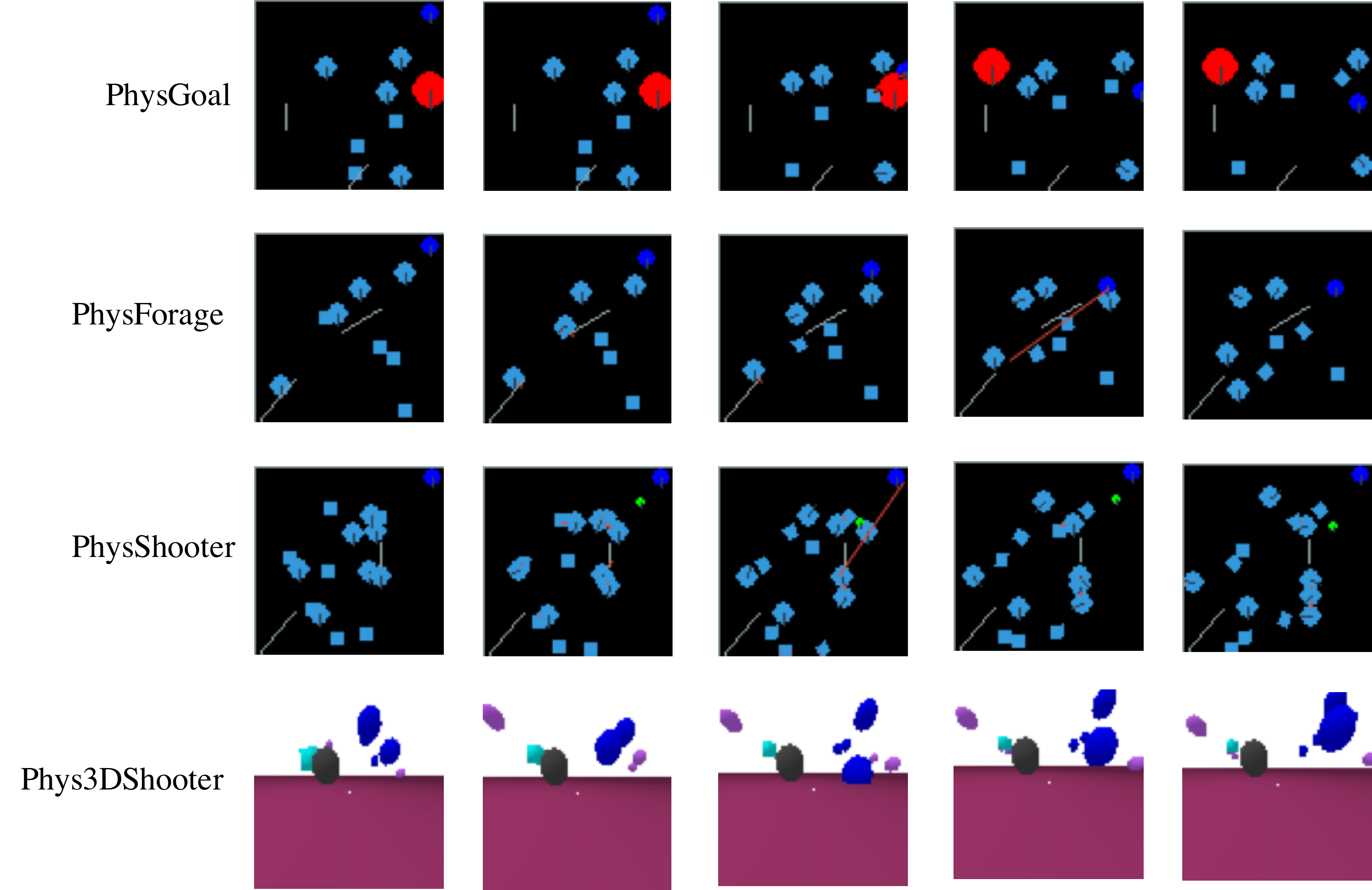}
\caption{\small Example agent game-play in each of the PhysWorld environments. In PhysGoal the dark blue agent attempts to reach a red goal while avoiding moving objects. In PhysForage the dark blue agent attempts to gather light blue circles while avoiding squares. In PhysShooter, the dark blue agent is immobile and chooses to fire bullet a green bullet at squares while avoiding circles. In Phys3DShooter, the grey fires turqoise bullets at purple spheres while avoiding blue spheres.} 
\label{fig:physworld_example}
\end{figure}

Examples of agents playing the PhysWorld environments are given in Figure ~\ref{fig:physworld_example}.

\subsection{\spatial Predictions} 
\begin{figure}
\centering
\includegraphics[width=0.9\linewidth]{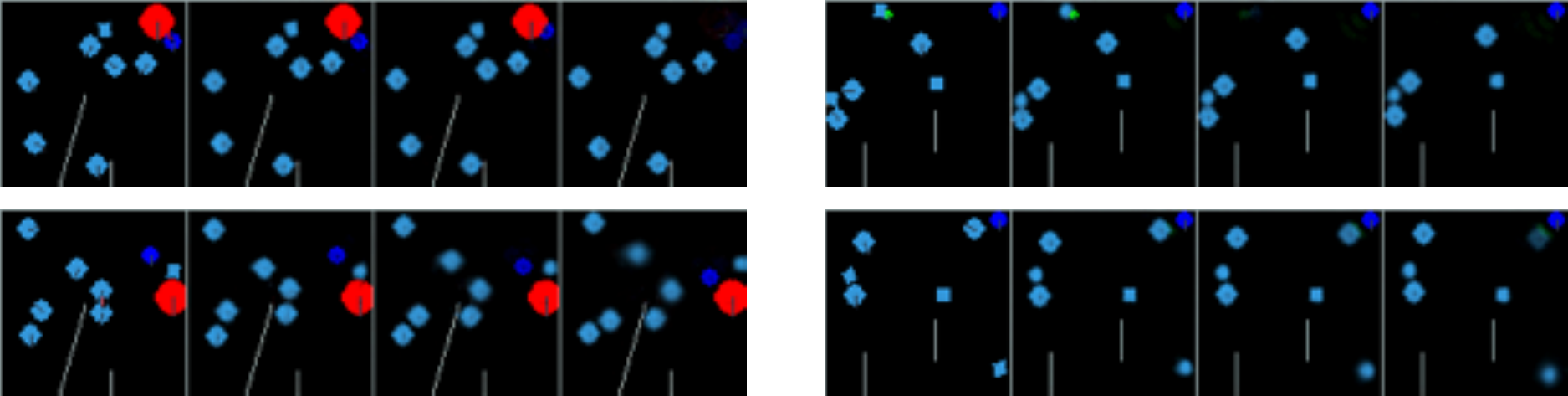}
\caption{\small Future image prediction on PhysGoal (left) and PhysShooter (right). First image is current observation, the next three are predicted. \spatial is able to predict future dynamics of boxes and balls and anticipate agent movement (PhysGoal) and agent shooting (PhysShooter).} 
\label{fig:physworld_qual}
\end{figure}
Figure \ref{fig:physworld_qual} shows the qualitative next 3 frame predictions of \spatial on each of the different PhysWorld environment with the first frame being the current observation. In PhysGoal, \spatial is able to infer the movement of the obstacles, the dark blue agent, and the red goal after agent collection. In PhysGather, \spatial is able to infer movement of obstacles as well as the gather of a circle. In PhysShooter, \spatial is able to anticipate a collision of the bullet with a moving obstacle and further infer the shooting of a green bullet by the agent.

\begin{figure}
\centering
\includegraphics[width=\linewidth]{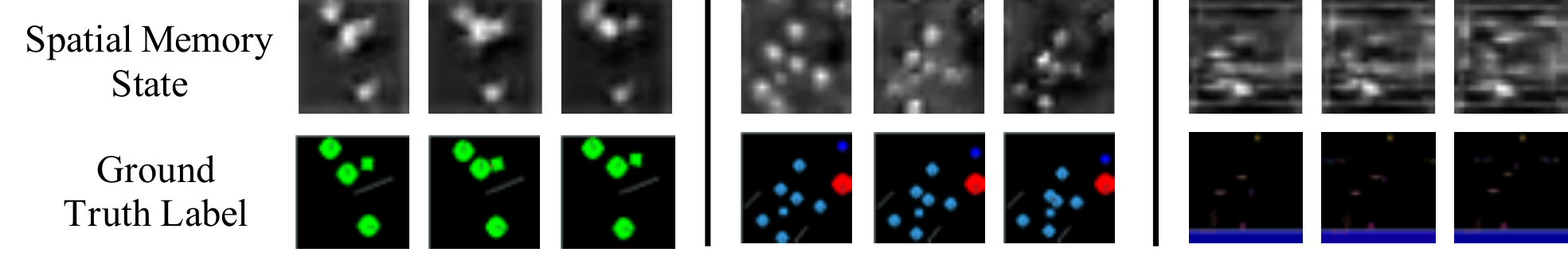}
\caption{\small Visualization of \spatial hidden state on PhysVideos (left), PhysGoal (middle) and Atari DemonAttack (right). Hidden state has high activations for moving objects while background objects such as walls (left), red goals (middle) and platforms (right) are not attended to as much.} 
\label{fig:spatial_mem}
\end{figure}

\subsection{Visualization of Spatial Memory} \label{sec:attention}
We provide visualization of the values of spatial memory hidden state while predicting future frames. We visualize the values of spatial memory on PhysVideos, PhysGoal and the Atari environment Demon Attack in Figure \ref{fig:spatial_mem}. To visualize, we take the mean across the channels of each grid pixel in the spatial memory hidden state. We find strong correspondence between high activation regions in the spatial memory and dynamic objects in the associated ground label of the dynamic objects. We further find that static background, such as walls in the input, goals and platforms appear to be passed along in input features.

\section{Additional Atari experiments}
\begin{figure*}[t]
\centering
\includegraphics[width=\linewidth]{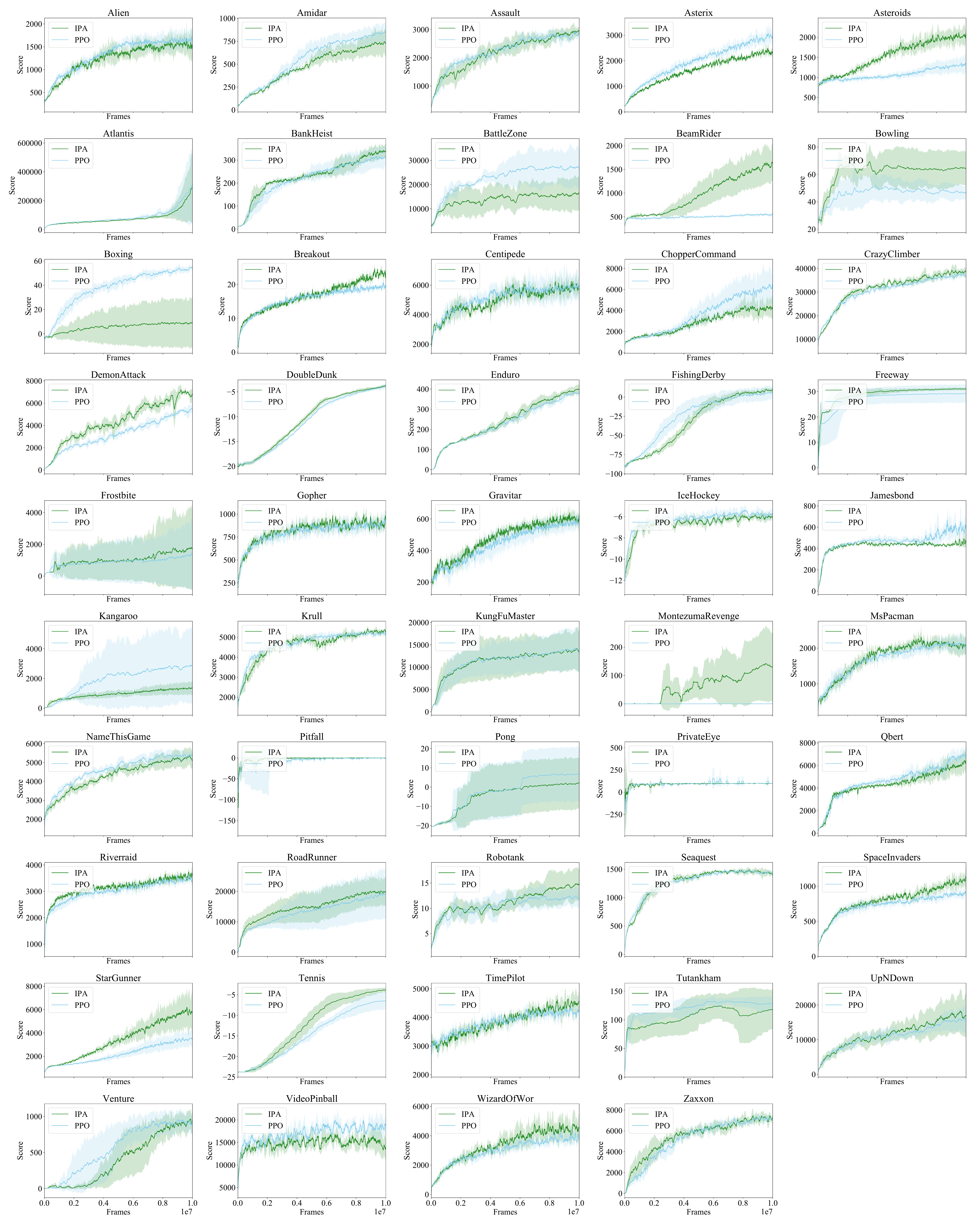}
\caption{\small Plots of policy performance trained with either PPO or \agent on all Atari environments on 5 different seeds. \agent sometimes leads to low learning early on the training due to rapid change of 3 predicted future frames. However, later on in training in many different environments, \agent provides performance gains by giving policies future trajectories.} 
\label{fig:atari_plot}
\end{figure*}
\begin{table*}[t]
\small
\centering
\begin{tabular}{ccc}
    \toprule
    Environment & PPO & D2A\\
    \midrule
    Alien & \textbf{ 1668.6 } $\pm$ 224.3 & 1485.5 $\pm$ 281.0 \\
    Amidar & \textbf{ 855.9 } $\pm$ 98.6 & 725.5 $\pm$ 135.0 \\
    Assault & 2939.2 $\pm$ 153.2 & \textbf{ 2968.4 } $\pm$ 124.0 \\
    Asterix & \textbf{ 2920.8 } $\pm$ 287.3 & 2334.0 $\pm$ 184.0 \\
    Asteroids & 1321.0 $\pm$ 233.5 & \textbf{ 2098.4 } $\pm$ 102.0 \\
    Atlantis & \textbf{ 323205.4 } $\pm$ 277643.2 & 289369.8 $\pm$ 239469.0 \\
    BankHeist & 310.4 $\pm$ 44.0 & \textbf{ 334.3 } $\pm$ 29.0 \\
    BattleZone & \textbf{ 26828.0 } $\pm$ 8472.0 & 16526.7 $\pm$ 6986.0 \\
    BeamRider & 553.1 $\pm$ 28.4 & \textbf{ 1630.3 } $\pm$ 400.0 \\
    Bowling & 46.6 $\pm$ 5.2 & \textbf{ 64.3 } $\pm$ 13.0 \\
    Boxing & \textbf{ 54.3 } $\pm$ 2.5 & 8.9 $\pm$ 20.0 \\
    Breakout & 19.7 $\pm$ 0.9 & \textbf{ 23.4 } $\pm$ 1.0 \\
    Centipede & \textbf{ 6043.7 } $\pm$ 990.6 & 6032.5 $\pm$ 199.0 \\
    ChopperCommand & \textbf{ 6549.4 } $\pm$ 1779.1 & 4112.0 $\pm$ 1024.0 \\
    CrazyClimber & 36893.2 $\pm$ 463.9 & \textbf{ 38499.0 } $\pm$ 1221.0 \\
    DemonAttack & 5510.9 $\pm$ 412.5 & \textbf{ 6793.6 } $\pm$ 558.0 \\
    DoubleDunk & -4.0 $\pm$ 0.5 & \textbf{ -3.8 } $\pm$ 0.0 \\
    Enduro & 376.7 $\pm$ 10.5 & \textbf{ 398.6 } $\pm$ 23.0 \\
    FishingDerby & 6.7 $\pm$ 10.1 & \textbf{ 9.3 } $\pm$ 3.0 \\
    Freeway & 29.2 $\pm$ 3.6 & \textbf{ 31.2 } $\pm$ 1.0 \\
    Frostbite & 1342.5 $\pm$ 2154.5 & \textbf{ 1701.1 } $\pm$ 2485.0 \\
    Gopher & 904.0 $\pm$ 42.3 & \textbf{ 941.1 } $\pm$ 56.0 \\
    Gravitar & 574.9 $\pm$ 36.2 & \textbf{ 627.2 } $\pm$ 25.0 \\
    IceHockey & \textbf{ -5.9 } $\pm$ 0.3 & -6.1 $\pm$ 0.0 \\
    Jamesbond & \textbf{ 598.9 } $\pm$ 112.1 & 454.3 $\pm$ 34.0 \\
    Kangaroo & \textbf{ 2842.4 } $\pm$ 2461.2 & 1373.0 $\pm$ 445.0 \\
    Krull & 5178.9 $\pm$ 205.1 & \textbf{ 5219.3 } $\pm$ 129.0 \\
    KungFuMaster & \textbf{ 13831.6 } $\pm$ 4483.6 & 13358.5 $\pm$ 4352.0 \\
    MontezumaRevenge & 0.0 $\pm$ 0.0 & \textbf{ 129.7 } $\pm$ 122.0 \\
    MsPacman & 1990.1 $\pm$ 227.9 & \textbf{ 2097.3 } $\pm$ 259.0 \\
    NameThisGame & \textbf{ 5406.4 } $\pm$ 278.0 & 5131.3 $\pm$ 427.0 \\
    Pitfall & -0.1 $\pm$ 0.3 & \textbf{ 0.0 } $\pm$ 0.0 \\
    Pong & \textbf{ 6.6 } $\pm$ 14.1 & 2.2 $\pm$ 13.0 \\
    PrivateEye & 95.6 $\pm$ 5.4 & \textbf{ 99.6 } $\pm$ 0.0 \\
    Qbert & \textbf{ 6981.0 } $\pm$ 548.0 & 6331.4 $\pm$ 769.0 \\
    Riverraid & 3411.0 $\pm$ 201.9 & \textbf{ 3612.4 } $\pm$ 130.0 \\
    RoadRunner & 19329.6 $\pm$ 8472.6 & \textbf{ 20041.8 } $\pm$ 4906.0 \\
    Robotank & 11.9 $\pm$ 1.8 & \textbf{ 14.9 } $\pm$ 3.0 \\
    Seaquest & \textbf{ 1426.0 } $\pm$ 43.5 & 1408.7 $\pm$ 51.0 \\
    SpaceInvaders & 902.4 $\pm$ 66.0 & \textbf{ 1132.6 } $\pm$ 101.0 \\
    StarGunner & 3450.0 $\pm$ 801.5 & \textbf{ 5778.5 } $\pm$ 1584.0 \\
    Tennis & -6.5 $\pm$ 2.1 & \textbf{ -3.8 } $\pm$ 1.0 \\
    TimePilot & 4281.8 $\pm$ 126.6 & \textbf{ 4580.0 } $\pm$ 314.0 \\
    Tutankham & \textbf{ 128.5 } $\pm$ 12.3 & 118.2 $\pm$ 35.0 \\
    UpNDown & 15872.3 $\pm$ 3995.3 & \textbf{ 16913.7 } $\pm$ 6344.0 \\
    Venture & 930.2 $\pm$ 137.9 & \textbf{ 946.7 } $\pm$ 167.0 \\
    VideoPinball & \textbf{ 18878.1 } $\pm$ 1251.7 & 13981.2 $\pm$ 2136.0 \\
    WizardOfWor & 3835.6 $\pm$ 404.7 & \textbf{ 4629.8 } $\pm$ 662.0 \\
    Zaxxon & 7197.4 $\pm$ 220.6 & \textbf{ 7271.0 } $\pm$ 264.0 \\
    \bottomrule
\end{tabular}
\caption{\small Scores obtained on Stochastic Atari Environments with \emph{sticky actions} (actions repeated with 50\% probability at each step). Scores are average performance over 100 episodes after 10M training frames, over 5 different random seeds.}
\label{tbl:atari_large_quantative}
\end{table*}

We provide plots of training curves on all Atari environments in Figure~\ref{fig:atari_plot} on provide on quantitative numbers in Figure~\ref{tbl:atari_large_quantative}.

\myparagraph{Predictions on Atari}
\begin{table*}
\centering
\begin{tabular}{lccc}
    \toprule
    Environment & MSE PD & MSE DN & Percent Advantage\\
    \midrule
    Assault & 0.00477 & 0.00522 & 9.4\%\\ 
    Asteroids & 0.002506 & 0.002518 & 4.7\% \\
    Breakout &  0.000417 & 0.000423 & 1.4\% \\
    DemonAttack & 0.00433 & 0.00562 & 29.8\% \\
    Enduro & 0.00576 & 0.00411 & -28.7\% \\
    FishingDerby & 0.00183 & 0.00192 & 4.9\% \\
    Frostbite & 0.000965 & 0.00107 & 10.8\% \\
    IceHockey & 0.000614 & 0.0013 & 111.7\% \\
    Pong & 0.00636 & 0.00584 & -8.2\% \\
    Tennis & 0.00142 & 0.00132 & -7.1\% \\
    \bottomrule
\end{tabular}
\vspace{5pt}
\caption{\small MSE on Stochastic Atari Environments (a action is repeated with a geometric distribution with p=0.5) at 1 million training frames. MSE PD is trained with a model from physics dataset while MSE DN is trained with a model from scratch. We evaluate percentage advantage for initializing with a physics dataset as compared to from scratch. We average 12.9\% decrease in MSE error using a initialization from pretraining on a physics dataset. The most negative environment, Enduro, involves a 3D landspace which initializing from model trained on a physics data set may be detrimental.}
\label{tbl:atari_transfer}
\end{table*}

We also investigate the benefits (in terms of MSE) of initializing \spatial pretrained on the physics dataset compared to training with scratch in Figure \ref{tbl:atari_transfer}. We evaluate the MSE error at 1 million frames and find that initializing with the physics dataset provides a 12.9\% decrease in MSE error. We find that pretraining helps on 7 of the 10 Atari environments, with the most negatively impacted environment being Enduro, a 3D racecar environment in which the environmental prior encoded by the physics dataset may be detrimental. More significant gains in transfer may be achievable by using a large online database of  2D YouTube videos which cover even more of diversity of games.

\begin{figure*}[t]
\centering
\includegraphics[width=\linewidth]{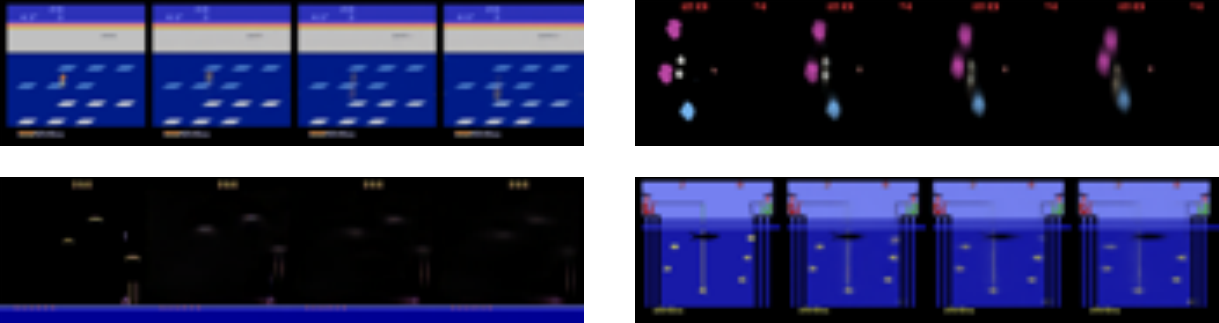}
\caption{\small Visualization of model future state prediction on 4 games in Atari (Frostbite - upper left, DemonAttack - lower left, Asteroids - upper right, FishingDerby - lower right). \spatial is able to predict falling of bullets, the catching of fish, movement of asteroids, and the movement of tiles/future agent movement in different environments. First frame visualized is ground truth observation, next 3 frames are model future frame predictions.} 
\label{fig:atari_qual}
\end{figure*}

\textbf{\spatial Predictions}
We further visualize qualitative results on \spatial on training Atari in Figure \ref{fig:atari_qual}. In general, across the Atari Suite, we found that \spatial is able to accurately model both the environment and agents behavior. In the figure, we seed that \spatial is able to accurately predict agent movement and ice block movement in Frostbite. On DemonAttack, \spatial is able to infer the falling of bullets. On Asteroid, \spatial is able to infer the movement of asteroids. Finally, on FishingDerby, \spatial is able to the right player capturing a fish and also predict that the left player is likely to catch a fish (indicated by the blurriness of the rod). We note that any blurriness in predicted output may in fact even be beneficial to the policy, since a policy can learn to interpret the input. 

\end{document}